\documentclass[letterpaper, 10 pt, conference]{ieeeconf}  

\IEEEoverridecommandlockouts                              

\usepackage{amsmath} 
\usepackage{amssymb}  
\usepackage{graphicx}
\usepackage{xcolor}
\usepackage{subcaption}
\usepackage{hyperref}
\usepackage{bm}
\usepackage{tikz}
\usetikzlibrary{shapes.geometric}
\usetikzlibrary{arrows.meta}
\usetikzlibrary{calc}
\usetikzlibrary{positioning}
\usetikzlibrary{decorations.pathreplacing}
\usetikzlibrary{tikzmark}

\usepackage{algorithm}
\usepackage{algorithmic}

\usepackage{printlen}

\usepackage{booktabs}

\usepackage{layouts}
\usepackage{units}
\usepackage{nicefrac, xfrac}
\usepackage{multirow}
 
\usepackage{array}
\usepackage{hyperref}
\usepackage{balance}
\newcolumntype{M}[1]{>{\centering\arraybackslash}m{#1}}

\definecolor{dark_green}{rgb}{0.18, 0.55, 0.35} 
\definecolor{dark_orange}{rgb}{0.8, 0.45, 0.0} 
\definecolor{greyish_blue}{rgb}{0.4, 0.6, 0.8} 

\definecolor{grey_background}{HTML}{EEEAEA}
\definecolor{rose_background}{HTML}{EEEAEA}

\uselengthunit{mm} 

\title{\LARGE \bf
Grasp Anything: Combining Teacher-Augmented Policy Gradient Learning with Instance Segmentation to Grasp Arbitrary Objects
}

\author{Malte Mosbach and Sven Behnke
\thanks{All authors are with the Autonomous Intelligent Systems group, 
	Computer Science Institute VI -- Intelligent Systems and Robotics -- and the Center for Robotics and the Lamarr Institute for Machine Learning and Artificial Intelligence, University of Bonn, Germany; \tt{mosbach@ais.uni-bonn.de}}
}

\newlength{\leftsubfigures}
\begin{document}

\maketitle
\thispagestyle{empty}
\pagestyle{empty}

\begin{abstract}
Interactive grasping from clutter, akin to human dexterity, is one of the longest-standing problems in robot learning. 
Challenges stem from the intricacies of visual perception, the demand for precise motor skills, and the complex interplay between the two.
In this work, we present Teacher-Augmented Policy Gradient (TAPG), a novel two-stage learning framework that synergizes reinforcement learning and policy distillation.
After training a teacher policy to master the motor control based on object pose information, TAPG facilitates guided, yet adaptive, learning of a sensorimotor policy, based on object segmentation.
We zero-shot transfer from simulation to a real robot by using Segment Anything Model for promptable object segmentation.
Our trained policies adeptly grasp a wide variety of objects from cluttered scenarios in simulation and the real world based on human-understandable prompts.
Furthermore, we show robust zero-shot transfer to novel objects. 
Videos of our experiments are available at \url{https://maltemosbach.github.io/grasp_anything}.
\end{abstract}

\section{Introduction}
\begin{figure}
        \centering
        \begin{subfigure}{0.245\columnwidth}
                \centering
                \includegraphics[width=\linewidth, height=2\linewidth]{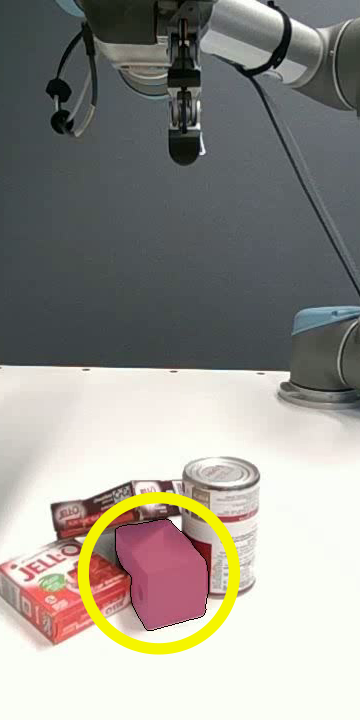}
                \footnotesize
                Target object to be grasped
            \end{subfigure}%
            \hfill
            \begin{subfigure}{0.245\columnwidth}
                \centering
                \includegraphics[width=\linewidth, height=2\linewidth]{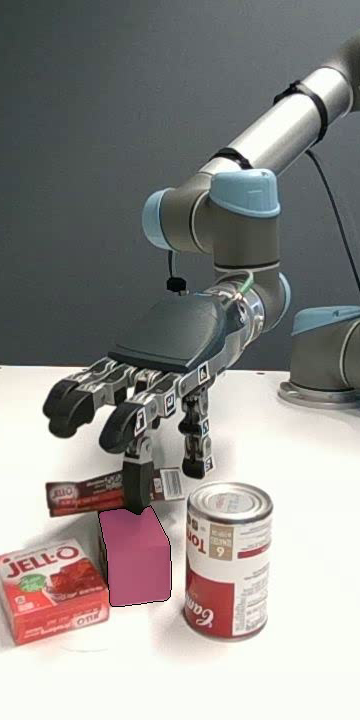}
                \footnotesize
                Reaching for target
            \end{subfigure}%
            \hfill
            \begin{subfigure}{0.245\columnwidth}
                \centering
                \includegraphics[width=\linewidth, height=2\linewidth]{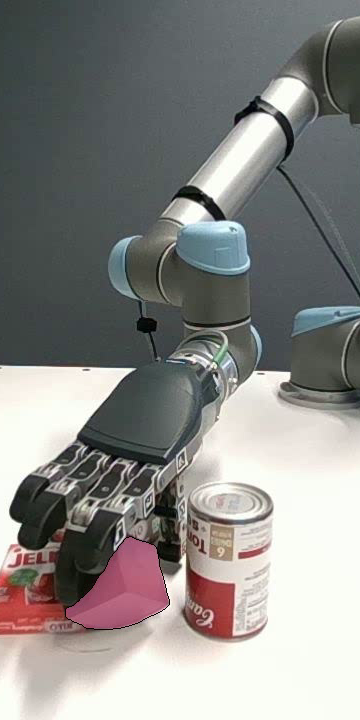}
                \footnotesize
                Interactions displace target
            \end{subfigure}%
            \hfill
            \begin{subfigure}{0.245\columnwidth}
                \centering
                \includegraphics[width=\linewidth, height=2\linewidth]{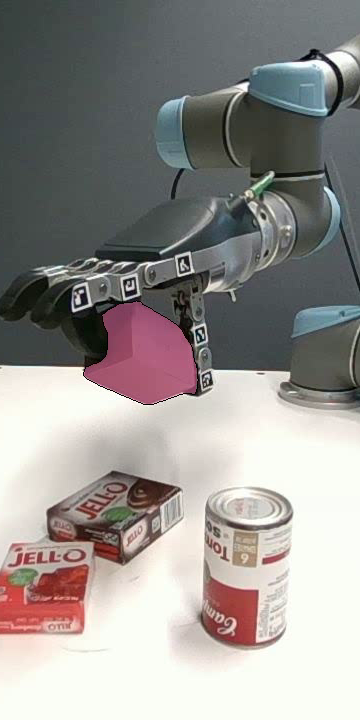}
                \footnotesize
                Successful retrieval
            \end{subfigure}%
        \caption{Robotic setup demonstrating the retrieval of a specific item from a cluttered environment.
        The agent dynamically grasps the user-specified object, recovering from failures and changes in the scene resulting from object interactions.}
        \label{fig:robot_setup}
\end{figure}

Grasping and retrieving a specific object is a fundamental, yet highly challenging sensorimotor skill, underpinning diverse manipulation tasks ranging from pick-and-place to tool-use.
While seemingly trivial for humans, grasping has been remarkably difficult to automate. 
Consider the setup in Figure~\ref{fig:robot_setup} where an anthropomorphic robot hand should perform grasping from a cluttered pile.
The target item is user-selected among multiple objects of unknown geometry.
Solving this task presents two key challenges: 
i) \textit{identifying and discerning targets} based on user cues and ii) the motor capability required for \textit{dexterous grasping and retrieval}.

The recent surge of vision foundation models (VFMs) has enabled semantic understanding and targeted segmentation of arbitrary scenes from human-understandable input. 
This can effectively address the identification of a target object, which can then be represented to a policy via its segmented point cloud.
However, while VFMs have found widespread adoption in computer vision, their perceptual power and understanding are rarely used in reinforcement learning (RL).

To understand this apparent disconnect, consider the prevalent training procedure of RL methods.
Using RL to learn sophisticated manipulation behaviors is notoriously sample-inefficient, requiring on the order of hundreds of millions of interactions---a problem that is compounded when learning from high-dimensional observations and in complex action spaces. Utilizing high-capacity vision models to produce observations pushes computational and sample complexities beyond currently feasible limits. 
To train visual policies more efficiently, policy distillation (PD) has recently been used to transfer knowledge from policies trained on low-dimensional states to high-dimensional visual observations~\cite{Chen2022,ChenTWKAA:SciRo23}.
However, differences between the characteristics of these observation spaces and resulting optimal behaviors have not yet been considered.
This neglect becomes particularly pronounced for tasks with substantial interplay between action and perception.
For targeted grasping, self-induced occlusion is the primary example of this phenomenon.
While the teacher learns to solve the task in an unrestricted manner, fully occluding an object during the grasping procedure may obstruct tracking and lead to missing information for the student policy.
Hence, different behaviors become feasible depending on the policy's observation-space -- a disparity not considered in prior works.

To address this limitation, we introduce a new guided student-teacher learning algorithm, Teacher-Augmented Policy Gradient (TAPG), that amalgamates ideas from RL and PD.
TAPG centers on the idea that the majority of the dexterous behaviors required to solve a manipulation task can be learned from low-dimensional observations, such that only the adaptation of this knowledge to a new observation space is required from the student.
As a result, only a small fraction of observations are generated in a high-dimensional, visual observation space.
This methodology enables the learning of perception-aware control strategies without the need to learn the entire policy from visual inputs.
We demonstrate this advantage by fine-tuning a vision-agnostic teacher with an additional target visibility reward. 
This enables the real-world deployment of the derived policies using a pre-trained VFM promptable object segmentation model.
To assess the robustness of the learned policies, we test them on unseen objects and in cluttered environments. 
These evaluations showcase strong zero-shot transfer and efficient recovery from failures.
In summary, our main contributions are:

\begin{itemize}
        \item We propose a novel two-stage learning framework, TAPG, that synergizes reinforcement learning and policy distillation to learn sensorimotor policies.
        \item We show how TAPG can be employed to integrate a pre-trained segmentation model with RL, resulting in a real-world executable, promptable grasping policy.
        \item We demonstrate the robustness of the learned behaviors in cluttered environments and on unseen objects.
\end{itemize}

\begin{figure*}
        \centering
        \begin{minipage}[b]{\leftsubfigures}
          \begin{subfigure}[b]{0.95\linewidth}
                \begin{tikzpicture}[]
                        \node[anchor=south west, inner sep=0] (privileged_state) at (0,0) {\includegraphics[width=.225\textwidth, height=.225\textwidth]{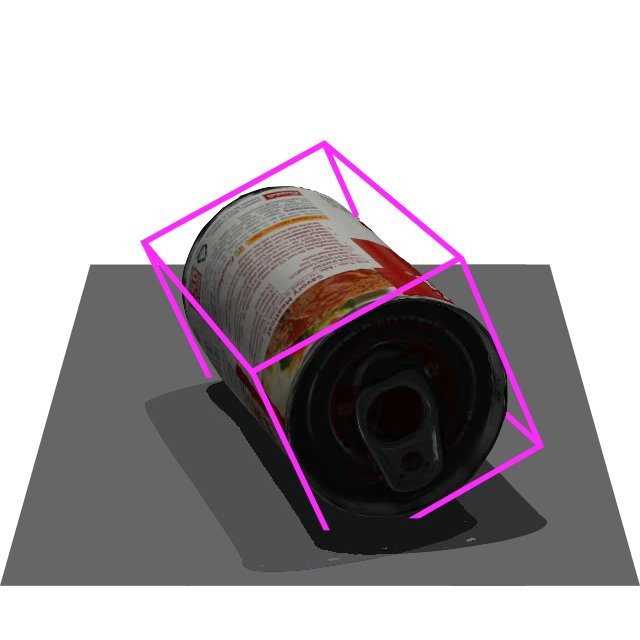}};
                        \node[below of=privileged_state, align=center, yshift=-0.2cm] (privileged_state_text) {privileged state $\bm{s}_t^\mathrm{T}$};
                        \begin{scope}[xshift=0.365\textwidth, yshift=1.7cm, scale=0.4, every node/.append style={transform shape}]
                                \newcommand{\inputnum}{2}
                                \newcommand{\hiddennum}{4}
                                \newcommand{\outputnum}{1}
                                \input{images/tikz_components/mlp.tex}
                        \end{scope}
                        \node[right of=privileged_state_text, xshift=0.205\textwidth, yshift=0.0cm, align=center] (policy_text){policy $\pi_\phi^\mathrm{T}$};

                        \node[anchor=south west, inner sep=0] (commands) at (5.75,0) {\includegraphics[width=.225\textwidth, height=.225\textwidth]{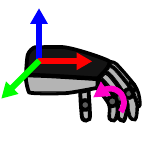}};
                        \node[below of=commands, align=center, yshift=-0.2cm] (commands_text) {actions $\bm{a}_t^\mathrm{T}$};

                        \draw[line width=0.3mm, -{Stealth[scale=1]}] (.225\textwidth,0.85cm) -- (.29\textwidth,0.85cm);

                        \node[xshift=0.315\textwidth, yshift=0.85cm] (priv_state) {$\bm{s}_t^\mathrm{T}$};
                        \node[xshift=0.315\textwidth, yshift=1.35cm] (previous_action) {$\bm{a}_{t-1}^\mathrm{T}$};

                        \draw[line width=0.3mm, -{Stealth[scale=1]}] (.475\textwidth,1.1cm) -- (.55\textwidth,1.1cm);

                        \draw[dashed, line width=0.3mm, color=gray, -{Stealth[scale=1]}] (.49\textwidth,1.1cm) -- (.49\textwidth,2.2cm) -- (.315\textwidth,2.2cm) -- (.315\textwidth,1.7cm);

                        \node[xshift=0.925\textwidth, yshift=1.1cm] (pg_loss) {$\mathcal{L}^\mathrm{PG}(\phi)$};
                        \draw[line width=0.3mm, -{Stealth[scale=1]}] (commands) -- (pg_loss);
                \end{tikzpicture}

            \caption{Stage 1 - Training of the teacher policy}
            \label{fig:overview_stage_1}
          \end{subfigure}\\

          \vspace{0.0cm}
          \begin{subfigure}[b]{0.95\linewidth}
                \begin{tikzpicture}[]
                        \node[anchor=south west, inner sep=0] (sensory_state) at (0,0) {\includegraphics[width=.225\textwidth, height=.225\textwidth]{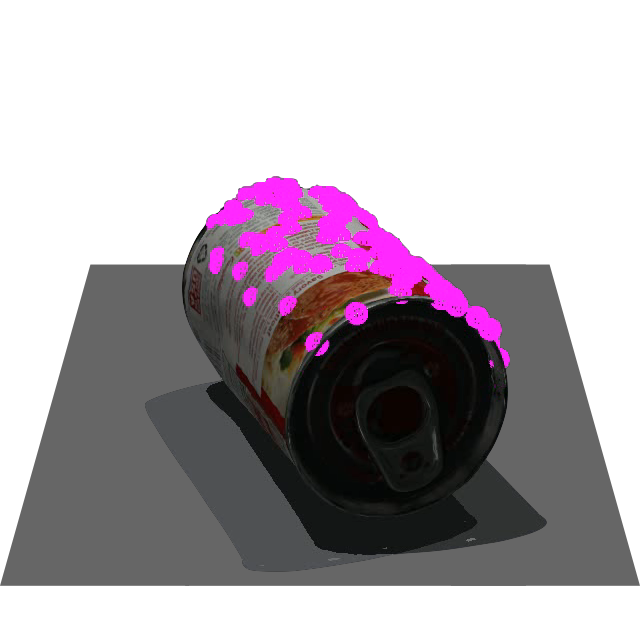}};
                        \node[below of=sensory_state, align=center, yshift=-0.2cm] (sensory_state_text) {sensory state $\bm{s}_t$};
                        \begin{scope}[xshift=0.365\textwidth, yshift=1.7cm, scale=0.4, every node/.append style={transform shape}]
                                \newcommand{\inputnum}{2}
                                \newcommand{\hiddennum}{4}
                                \newcommand{\outputnum}{1}
                                \input{images/tikz_components/mlp.tex}
                        \end{scope}
                        \node[right of=privileged_state_text, xshift=0.205\textwidth, yshift=0.0cm, align=center] (policy_text){policy $\pi_\theta$};

                        \node[align=center] (commands) at (.68\textwidth, 1.1cm) {actions $\bm{a}_t$};

                        \draw[line width=0.3mm, -{Stealth[scale=1]}] (.225\textwidth,0.85cm) -- (.29\textwidth,0.85cm);

                        \node[xshift=0.315\textwidth, yshift=0.85cm] (state) {$\bm{s}_t$};
                        \node[xshift=0.315\textwidth, yshift=1.35cm] (previous_action) {$\bm{a}_{t-1}$};

                        \draw[line width=0.3mm, -{Stealth[scale=1]}] (.475\textwidth,1.1cm) -- (.55\textwidth,1.1cm);

                        \draw[dashed, line width=0.3mm, color=gray, -{Stealth[scale=1]}] (.49\textwidth,1.1cm) -- (.49\textwidth,2.2cm) -- (.315\textwidth,2.2cm) -- (.315\textwidth,1.6cm);

                        \node[xshift=0.925\textwidth, yshift=1.1cm] (pg_loss) {$\mathcal{L}^\mathrm{PG}(\theta)$};
                        \draw[line width=0.3mm, -{Stealth[scale=1]}] (commands) -- (pg_loss);
                        
                        \node[above of=commands, align=center, yshift=0.18cm] (bc_loss) {$\mathcal{L}^\mathrm{BC}(\theta, \phi)$};
                        \draw[line width=0.3mm, -{Stealth[scale=1]}] (commands) -- (bc_loss);

                        \node[above of=commands, align=center, yshift=1.5cm] (teacher_action) {$\bm{a}_{t}^\mathrm{T}$};
                        \draw[line width=0.3mm, -{Stealth[scale=1]}] (teacher_action) -- (bc_loss);

                        \draw[line width=0.3mm] (5.5cm + .1125\textwidth, 2.9cm) -- (6.0cm + .1125\textwidth, 3.1cm);
                        \draw[line width=0.3mm] (5.5cm + .1125\textwidth, 3.0cm) -- (6.0cm + .1125\textwidth, 3.2cm);
                        \node[xshift=7.125cm + .1125\textwidth, yshift=3.1cm] (sg) {stop gradient};
                \end{tikzpicture}
                
            \caption{Stage 2 - Teacher-guided training of the sensorimotor policy}
            \label{fig:overview_stage_2}
          \end{subfigure}
        \end{minipage}%
        \begin{minipage}[b]{\dimexpr\textwidth-\leftsubfigures}
          \begin{subfigure}[b]{0.95\linewidth}
            \begin{tikzpicture}
                \node[rotate=90, align=center, font={\small}, xshift=0.1125\textwidth] (tracker_segmentation_text) at (0, 0.3) {tracking};
                \node[anchor=south west, inner sep=0] (rollout_seg_1) at (0.3,0.3) {\includegraphics[width=.225\textwidth]{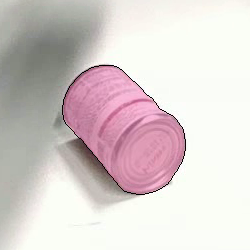}};
                \node[anchor=south west, inner sep=0, right=0.1cm of rollout_seg_1] (rollout_seg_2) {\includegraphics[width=.225\textwidth]{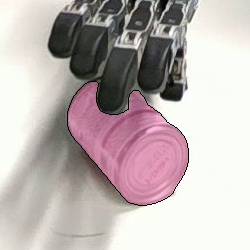}};
                \node[anchor=south west, inner sep=0, right=0.1cm of rollout_seg_2] (rollout_seg_3) {\includegraphics[width=.225\textwidth]{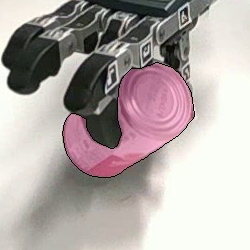}};
                \node[anchor=south west, inner sep=0, right=0.1cm of rollout_seg_3] (rollout_seg_4) {\includegraphics[width=.225\textwidth]{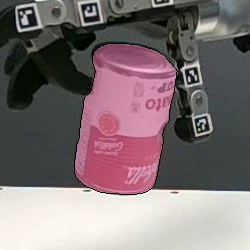}};
                
                \node[rotate=90, align=center, font={\small}, xshift=0.1125\textwidth] (rollout_text) at (0, -1.4) {rollout};
                \node[anchor=south west, inner sep=0] (rollout_1) at (0.3,-1.4) {\includegraphics[width=.225\textwidth]{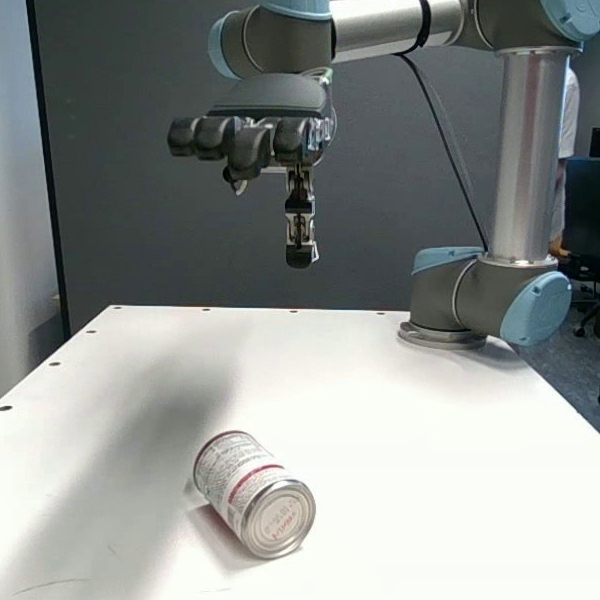}};
                \node[anchor=south west, inner sep=0, right=0.1cm of rollout_1] (rollout_2) {\includegraphics[width=.225\textwidth]{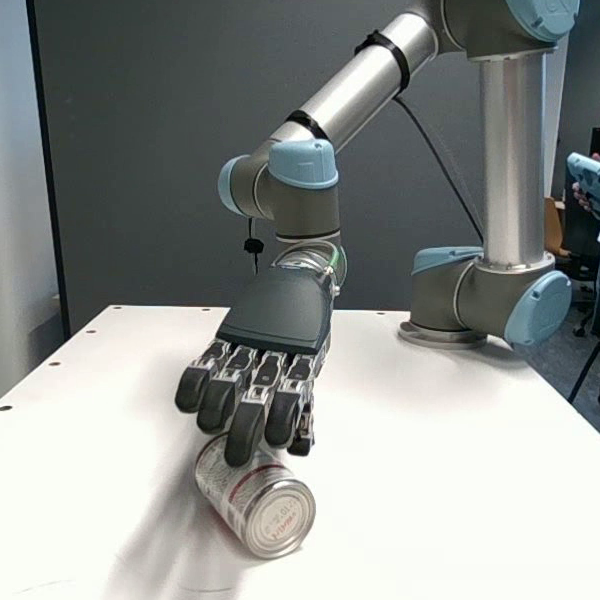}};
                \node[anchor=south west, inner sep=0, right=0.1cm of rollout_2] (rollout_3) {\includegraphics[width=.225\textwidth]{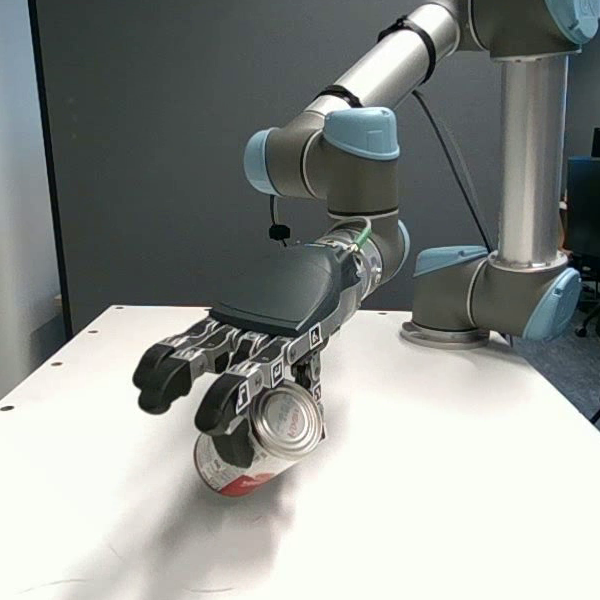}};
                \node[anchor=south west, inner sep=0, right=0.1cm of rollout_3] (rollout_4) {\includegraphics[width=.225\textwidth]{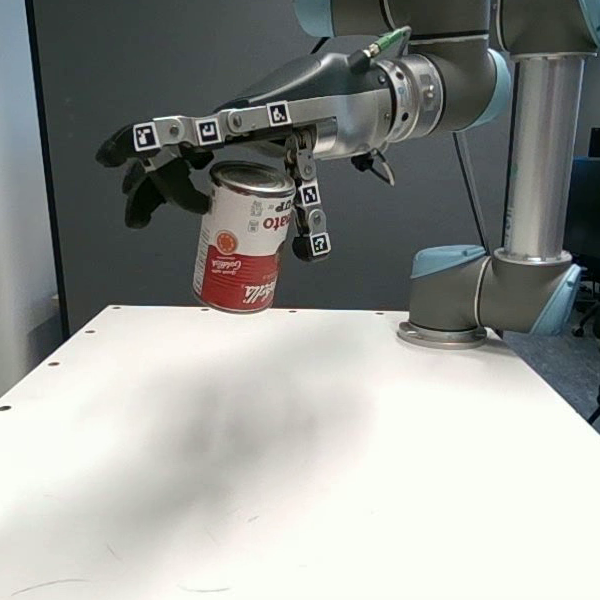}};

                \node[draw, 
                        fill=blue!20, 
                        text width=5cm, 
                        align=center, 
                        minimum height=0.8cm, 
                        rounded corners, 
                        anchor=south west,
                ] 
                (sam) at (0, 2.5) 
                {Segment Anything Model};

                \node[anchor=south west] (sam_initializes) at (.1125\textwidth + 0.3cm, 2.0) {initialize};

                \draw[line width=0.3mm, -{Stealth[scale=1]}] (0.3,-1.85) -- (\textwidth,-1.85) node[anchor=south east, font={\small}] {$T$};
                \node[below of=rollout_1, align=center, font={\small}, yshift=0.025cm] {$t=0$};

                \node[anchor=south west, inner sep=0] (camera_image) at (0,4.0) {\includegraphics[width=.25\textwidth]{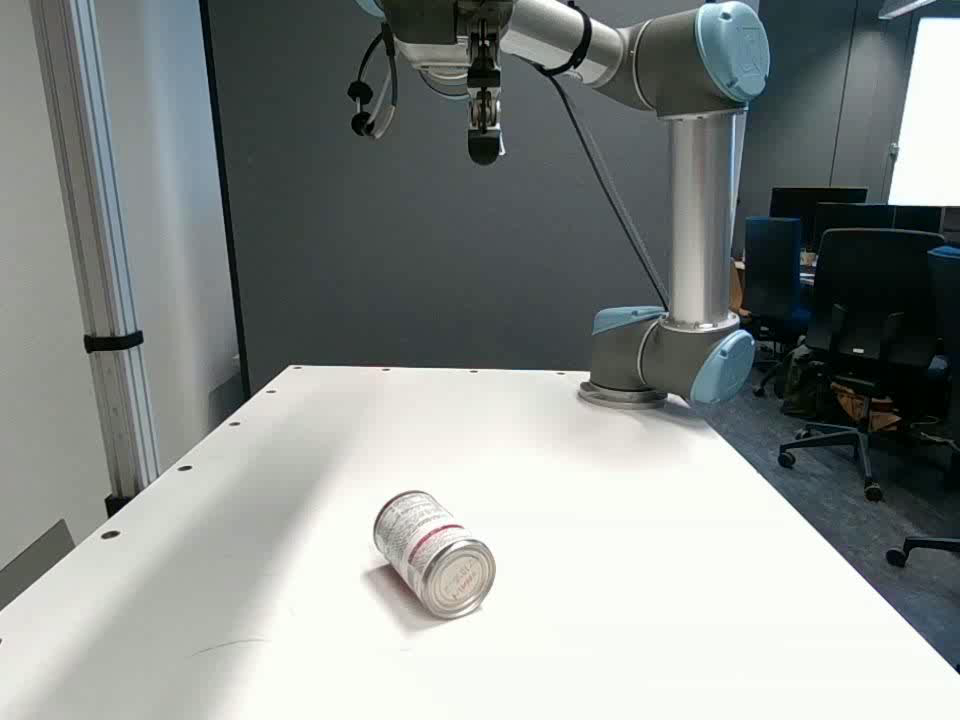}};
                \node[anchor=south west, inner sep=0] (point_prompt) at (2.5,4.0) {\includegraphics[width=0.1875\textwidth, height=0.1875\textwidth]{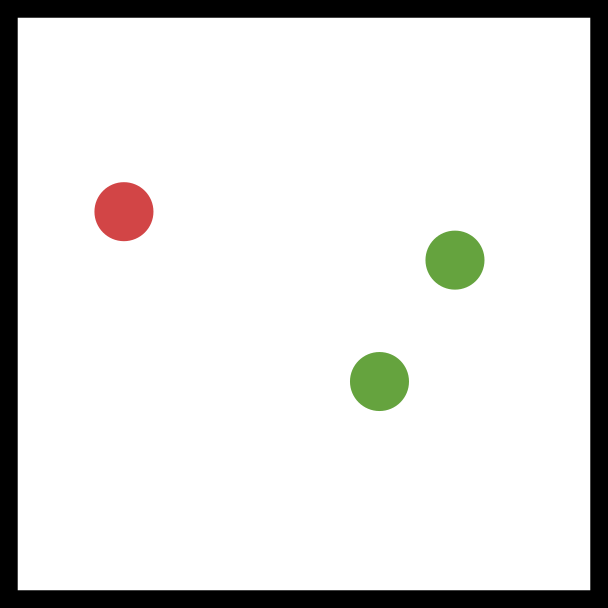}};
                \node[anchor=south west, inner sep=0] (text_prompt) at (4.,4.0) {\includegraphics[width=0.1875\textwidth, height=0.1875\textwidth]{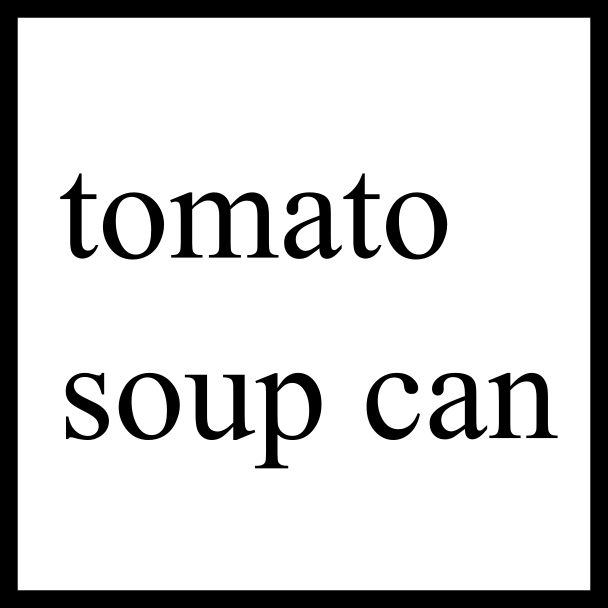}};

                \node[above of=camera_image, align=center, yshift=-0.1cm] (text) {image};
                \node[above of=point_prompt, align=center, yshift=-0.1cm, xshift=0.8cm] (text2) {target prompt};

                \draw[decorate,decoration={brace,amplitude=8pt,mirror}, line width=0.3mm] (2.51,4.0) -- (4.0cm + 0.1875\textwidth,4.0) node[midway,below=5pt] {};
                \draw[line width=0.3mm, -{Stealth[scale=1]}] (3.887, 3.75) -- (3.887, 3.3);
                \draw[line width=0.3mm, -{Stealth[scale=1]}] (camera_image.south) -- (0.125\textwidth, 3.3);
                \draw[line width=0.3mm, -{Stealth[scale=1]}] (.1125\textwidth + 0.3cm, 2.5) -- (rollout_seg_1.north);

            \end{tikzpicture}
            \caption{Real-world deployment of promptable grasping policy}
            \label{fig:overview_deployment}
          \end{subfigure}
        \end{minipage}
        \caption{Overview of the proposed two-stage learning framework. 
        In the first stage (Fig.~\ref{fig:overview_stage_1}), a teacher policy $\pi_\phi^\mathrm{T}$ is trained to solve the grasping task from privileged state information $\bm{s}_t^\mathrm{T}$. 
        Here, the goal is to frame the problem as a tractable RL task. 
        In the second stage (Fig.~\ref{fig:overview_stage_2}), the proficient teacher is used to guide the training of the sensorimotor policy $\pi_\theta$. 
        Replacing ground-truth segmentations with the output of a promptable segmentation model enables real world deployment (Fig.~\ref{fig:overview_deployment}).}
        \label{fig:overview}
\end{figure*}

\section{Related Work}
\subsection{Vision Foundation Models}
Vision foundation models (VFMs), high-capacity deep learning models pre-trained on vast datasets, have become a cornerstone of computer vision (CV) research over the past decade.
The early success of convolutional neural networks trained on visual datasets like ImageNet demonstrated the effectiveness of this paradigm~\cite{Krizhevsky2012}.
Motivated by the transformative success of promptable large language models, there has been renewed interest to mirror this paradigm in CV.
This has led to the emergence of powerful VFMs, such as the Segment-Anything-Model (SAM)~\cite{Kirillov2023}, X-Decoder~\cite{Zou2023}, and SegGPT~\cite{Wang2023}.
SAM is an image segmentation model that can be conditioned with flexible prompts such as points or bounding boxes to segment arbitrary objects in images and has recently been extended to track segmentations in videos~\cite{Cheng2023}.
This provides a robust foundation for identifying and representing a target object to be grasped without limiting the scope of application to specific object sets.

\subsection{Learning-based Robotic Grasping}
The task of robotic grasping has been studied extensively in the literature, with particular challenges arising from unknown object geometries and cluttered environments. 
The advent of deep RL has brought forth systems that learn to grasp objects through trial and error.
Pioneering work by Levine et al.~\cite{Levine2018} employs deep and reinforcement learning approaches to grasp objects from clutter, resulting in a continuous control policy. 
Building on this foundation, Kalashnikov et al.~\cite{Kalashnikov2018} used real-world and simulated data to train large-scale policies for grasping, outperforming prior methods.
Zeng et al.~\cite{Zeng2018} use deep Q-learning to learn synergies between pre-grasp manipulation (e.g. pushing) and grasping behaviors to retrieve specific objects from clutter.
Another significant line of work focuses on combining RL with human demonstrations. 
Vecerik et al.~\cite{Vecerik2017} demonstrated the successful integration of human demonstrations and deep RL to accelerate training and enhance performance. 
Similarly, Rajeswaran et al.~\cite{Rajeswaran2018} proposed a method to incorporate demonstrations into policy gradient methods, accelerating progress on manipulation tasks with sparse rewards. Behnke and Pavlichenko~\cite{PavlichenkoB:CASE23} designed a dense multi-component reward function for learning dexterous pre-grasp manipulation and functional 
grasping of novel instances of object categories with an anthropomorphic hand.
Quillen et al.~\cite{Quillen2018} introduced an approach that uses Q-learning with deep neural networks applied to grasping in clutter. 
This work emphasized real world scenarios and simulation-to-real transfer, highlighting the challenges and solutions related to cluttered environments.

\subsection{Integrating Imitation with Reinforcement Learning}
The integration of imitation learning (IL) objectives with RL has emerged as a powerful strategy. 
Two principal paradigms can be distinguished: imitating expert demonstrations and imitating another policy.
Firstly, expert demonstrations can be used to streamline the exploration process in RL.
Specifically, demonstrations can be employed to initialize a policy~\cite{Rajeswaran2018}, or guide the learning process of on-policy~\cite{Rajeswaran2018} or off-policy~\cite{Vecerik2017,Nair2018a} methods.
Rajeswaran et al.~\cite{Rajeswaran2018} propose a system that combines ideas from RL and behavior cloning (BC)~\cite{Pomerleau1988}.
Their work demonstrates how adding an objective that maximizes the probability of the expert actions can be used to foster the learning process of policy gradient methods on dexterous manipulation tasks.
Mosbach and Behnke learn the use of tools from a single demonstration by employing non-rigid grasp-pose registration in a shape space~\cite{MosbachB:CASE23}.

Secondly, imitation can be used to transfer knowledge between policies, also known as policy distillation (PD)~\cite{Czarnecki2019}.
A specific formulation of this framework has recently proven useful in the context of robot learning~\cite{Chen2022, ChenTWKAA:SciRo23}, wherein a teacher policy is trained from privileged, low-dimensional observations and then distilled into a visual student policy.
The distillation from teacher to student is performed using DAgger~\cite{Ross2011}.
Our proposed method amalgamates ideas from both RL and PD paradigms, facilitating knowledge transfer between observation spaces, while enabling unrestricted learning of the student policy.

\section{Method}

In sensory-motor robot learning, a key challenge is to devise a continuous control policy that can interpret sensory observations and act on them.
Formally, we consider the problem of learning a policy, $\pi_\theta$, which, given high-dimensional sensory observations $\bm{s}_t$, yields actions $\bm{a}_t$ that control a robot.
The objective is to adjust the parameter set $\theta$ to maximize the discounted sum of rewards
\begin{equation}
        J(\pi_\theta) = \underset{\bm{s}_t, \bm{a}_t \sim \rho_\pi}{\mathbb{E}} \left[ \gamma^t r_t \right],
\end{equation}
where $\gamma$ is a discount factor and $\rho_\pi$ denotes the joint distribution of state-action pairs under the policy $\pi$.
The rewards $r_t$ encapsulate task-objectives like grasping an object.

This direct formulation of sensorimotor learning intertwines two difficult problems: \textit{deriving underlying state-representations} from high-dimensional data and \textit{determining optimal behavior} in each state \cite {Chen2020}.
Fortunately, student-teacher learning has been shown to efficiently disentangle both tasks~\cite{Chen2020, Chen2022}.
Initially, a teacher policy, $\pi^\mathrm{T}_\phi$, learns to solve the manipulation task given privileged access to the environment's low-dimensional state  $\bm{s}_t^\mathrm{T}$.
The succinct nature of these inputs makes learning significantly more sample-effective than learning from images.
In synergy, low-dimensional observations can be generated efficiently in massively parallelized simulators~\cite{Makoviychuk2021}, providing abundant training data.
The trained teacher policy offers ample supervision for the learning of the sensorimotor policy.
We collect demonstrations (state-action tuples) from the teacher policy in a dataset, denoted $\mathcal{D}^T = \{ (\bm{s}_i, \bm{a}_i^T) \}$.
Our method bridges the gap between imitation and reinforcement learning -- a combination that allows for sensorimotor policies to adapt to the nuances of visual perception.

\subsection{Teacher-Augmented Policy Gradient (TAPG)}

Policy gradient (PG) methods center on the idea of increasing the probabilities of actions that yield higher returns, and decreasing the probabilities of actions that lead to lower returns.
Therefore, an estimate of the policy gradient is optimized with stochastic gradient ascent. 
Most commonly, a gradient estimator of the form
\begin{equation}
        \hat{g} = \underset{\bm{s}_t, \bm{a}_t \sim \rho_\pi}{\hat{\mathbb{E}}_t} \left[ \nabla_\theta \log \pi_\theta (\bm{a}_t | \bm{s}_t) \hat{A}^{\pi_\theta}(\bm{s}_t, \bm{a}_t) \right]
\end{equation}
is employed, where $\hat{\mathbb{E}}_t[\cdot]$ represents the mean over a finite batch of samples and $\hat{A}$ is an estimate of the advantage function~\cite{Schulman2017}.
The corresponding loss function is given by
\begin{equation}
        \mathcal{L}^\mathrm{PG}(\theta) = -\underset{\bm{s}_t, \bm{a}_t \sim \rho_\pi}{\hat{\mathbb{E}}_t} \left[ \log \pi_\theta (\bm{a}_t | \bm{s}_t) \hat{A}^{\pi_\theta}(\bm{s}_t, \bm{a}_t) \right].
\end{equation}
Multiple methods have been proposed that take constrained updates on this objective to avoid policy collapse~\cite{Schulman2017, Schulman2015}.

Although PG approaches have achieved impressive results in robot learning from low-dimensional states, they generally require a vast number of samples.
This problem is exacerbated in high-dimensional settings.
Tackling complex sensorimotor manipulation tasks necessitates guidance, i.e., from demonstrations or an adept teacher policy.
Cloning the behaviors from a teacher policy or dataset (BC) corresponds to solving the following maximum likelihood problem:
\begin{equation}
        \underset{\theta}{\max}\underset{\bm{s}_t, \bm{a}_t \sim \mathcal{D}^\mathrm{T}}{\hat{\mathbb{E}}_t} \left[ \log \pi_\theta (\bm{a}_t | \bm{s}_t)\right].
\end{equation}
DAgger~\cite{Ross2011} improves the performance guarantees of BC by iteratively collecting new data from the current policy and adding it to the dataset.
However, the teacher policy is assumed to be optimal.
Given that student-teacher learning in RL is fundamentally concerned with transferring knowledge between different observation spaces, it is desirable for the student to adjust its behavior in response to novel information or challenges that arise from its expressive observations.

To formulate a guiding yet non-restrictive imitation objective, we make use of the fact that the teacher policy has already learned a value function, $V^{\pi^\mathrm{T}_\phi}(\bm{s}_t^\mathrm{T})$.
Using this additional knowledge, we introduce a gating term that estimates when to trust the teacher:
\begin{equation}
        \hat{A}^\mathrm{T} = [ V^{\pi^\mathrm{T}_\phi}(\bm{s}_t^\mathrm{T}) - V^{\pi_\theta}(\bm{s}_t) ]_{>0},
\end{equation}
where $[\cdot ]_{>0}$ evaluates to 1 if the argument is positive, and to 0 otherwise.
Integrating this term into the BC objective yields the following loss function:
\begin{equation}
        \mathcal{L}^\mathrm{BC}(\theta, \phi) = -\underset{\bm{s}_t, \bm{a}_t \sim \mathcal{D}^\mathrm{T}}{\hat{\mathbb{E}}_t} \left[ \log \pi_\theta (\bm{a}_t | \bm{s}_t)\hat{A}^\mathrm{T} \right],
\end{equation}
which incentivizes the student policy to imitate the teacher policy only if the teacher is estimated to be better in a given state.
Considering the relation
\begin{equation}
        \begin{split}
        \mathcal{L}^{\mathrm{BC}}(\theta, \phi) & = -\underset{\bm{s}, \bm{a}^\mathrm{T} \sim \mathcal{\pi^\mathrm{T}_\phi}}{\mathbb{E}} \left[ \log \pi_\theta (\bm{a}^\mathrm{T} | \bm{s}) A^\mathrm{T} \right] \\
        & = \mathrm{H}^{\times}(\pi_\phi^\mathrm{T}(\bm{s}) || \pi_\theta(\bm{s})) A^\mathrm{T},
        \end{split}
\end{equation}
it can be shown that for a sufficiently small imitation loss ($\mathcal{L}^{\mathrm{BC}} \le \mathrm{H}(\pi_\phi^\mathrm{T}(\bm{s})) \forall \bm{s} \in \mathcal{S}$), the student policy is guaranteed to be at least as good as the teacher, i.e., $\underset{\bm{s} \sim \mathcal{S}_0}{\mathbb{E}} [V^{\pi_\theta}(\bm{s})] \ge \underset{\bm{s} \sim \mathcal{S}_0}{\mathbb{E}} [V^{\pi_\phi^\mathrm{T}}(\bm{s})]$.
The full proof can be found in~\cite{Czarnecki2019}.

The resulting method, which we refer to as teacher-augmented policy gradient (TAPG), optimizes the combined objective $\mathcal{L}^\mathrm{PG}(\theta) + \mathcal{L}^\mathrm{BC}(\theta, \phi)$. 
The training procedure is outlined in Algorithm \ref{alg:tapg}.

\begin{algorithm}[t]
        \caption{Teacher-Augmented Policy Gradient (TAPG)}
        \label{alg:tapg}
        \begin{algorithmic}[1] 
            \REQUIRE teacher policy $\pi_\phi^\mathrm{T}$, empty teacher dataset $\mathcal{D}^\mathrm{T}$
            \FOR{$k = 0$ to $N$}
            \STATE Collect trajectories $\mathcal{D}_k$ by running the current policy $\pi_k = \pi(\theta_k)$.
            \STATE Compute policy advantage estimates $\hat{A}^{\pi_\theta}$.
            \STATE Compute teacher advantage estimates $\hat{A}^\mathrm{T}_t$.
            \STATE Update parameters $\theta_k$ by optimizing the combined loss $L^{PG}(\theta_k) + L^{BC}(\theta_k, \phi)$.
            \ENDFOR
        \end{algorithmic} 
\end{algorithm}

For the task at hand, we instantiate TAPG as outlined in Figure \ref{fig:overview}. 
In the first stage, we train the teacher policy to grasp objects represented by their oriented 3D bounding boxes. 
This representation is low-dimensional to allow for efficient learning while conveying enough information for specialized, geometry-aware behaviors to emerge~\cite{Mosbach2022}.
We subsequently learn a student policy from segmented point clouds of the object to be grasped.
Although operating in a guided learning setting, generating the amounts of data required to imitate a teacher policy or adapt its behaviors with RL is impractical with a VFM in the loop.
Instead, we utilize the ground truth segmentations provided by the simulator.
In the ideal case, the segmentation model will output identical masks in deployment.
To nevertheless account for the fact that tracked segmentations are less robust to high degrees of occlusion, we introduce an auxiliary reward term that encourages behaviors benign to the tracking model by maximizing the visibility of the target object.

\subsection{Perception-driven Reward Design}
Avoiding difficulties in object tracking amounts to avoiding large occlusions to the target object.
Therefore, in addition to task objectives, we optimize for behaviors that retain the visibility of the target object, as quantified by the visibility ratio $r_v$:
defined as an object's visible area over its total area in the image plane.
To estimate $r_v$, we project points sampled on the object's surface to the image plane of our camera.
The visibility ratio is then estimated as the number of projected points that fall into the object's segmentation mask over the total number of points.

\subsection{Sim-To-Real Adaptation}
A major impediment in sim-to-real deployment is the discrepancy between the simulated and real-world dynamics of the robot.
While the robot arm reaches the desired targets precisely in both settings, the anthropomorphic hand presents intricate joint couplings due to its tendon-actuated control.
To accurately emulate the hand's dynamics, we perform a calibration process where the fingers are opened and closed at various velocities.
Since the used Schunk SIH hand does not have joint encoders, we attach ArUco markers to the finger phalanges (the markers can be seen in Figure~\ref{fig:robot_setup} and are only used during calibration), allowing for angle tracking via a camera.
Subsequently, we fitted the simulated robot's response curves through a least squares polynomial approximation to match the real-world data.

Robotic manipulation inherently necessitates contact with the environment. 
However, manipulation should be performed exclusively with the hand, while the arm should not collide with the environment. 
To encourage this behavior, we add a term to the reward function that penalizes the policy for applying excessive contact forces to the robot arm.
\begin{table}[t]
        \caption{Rewards combine terms for \textcolor{dark_orange}{\bf task completion}, \textcolor{dark_green}{\bf directed exploration}, and \textcolor{greyish_blue}{\bf safety}. 
        The sensorimotor policy is additionally rewarded for ease of perception.}

        \begin{minipage}[b]{1 \linewidth}
                \centering 
         {\small       \begin{tabular}{llr}
                        \toprule
                        Term & Equation & Weight \\
                        \midrule
                        \tikzmarknode{task_start}{Sparse task reward} & $[\lVert \Delta \bm{p}_t \rVert _2 ]_{< \overline{p}}$ & 50.0  \\
                        Dense task reward & $\displaystyle  1 / (\lVert \Delta \bm{p}_t \rVert _2 + \epsilon_p)$ & \tikzmarknode{task_end}{1.0} \\
                        \addlinespace[1.1mm]
                        \tikzmarknode{exploration_start}{Fingertips close to object} & $\displaystyle  \sum_{i=1}^{F} \lVert \bm{p}_t^{f_i} - \bm{p}_t^{o} \rVert _2^2$ & 0.1  \\
                        Clearance reward & $\displaystyle 1 / (| \Delta h | + \epsilon_h)$ & \tikzmarknode{exploration_end}{1.0}  \\
                        \addlinespace[1.6mm]
                        \tikzmarknode{safety_start}{Large actions penalty} & $\displaystyle \lVert \bm{a}_t \rVert_2^2$ & -0.01  \\
                        Contact penalty & $\displaystyle [\lVert \bm{c}_t \rVert _2 ]_{> \overline{c}}$ & \tikzmarknode{safety_end}{1.0}  \\
                        \midrule
                        Target visibility reward & $\displaystyle r_v$ & {20.0}  \\
                        \bottomrule
                \end{tabular}
					}						
                \label{tab:reward_terms}
        \end{minipage}
        				
        \begin{tikzpicture}[overlay, remember picture, line width=1pt, rounded corners=2pt, line cap=round]
          \draw[dashed, dark_orange] ($(task_start.north west)+(-2mm,1mm)$) rectangle ($(task_end.south east)+(2mm,-1.2mm)$);
          \draw[dashed, dark_green] ($(exploration_start.north west)+(-2mm,3.5mm)$) rectangle ($(exploration_end.south east)+(2mm,-1.2mm)$);
          \draw[dashed, greyish_blue] ($(safety_start.north west)+(-2mm,1mm)$) rectangle ($(safety_end.south east)+(2mm,-1.2mm)$);
          \end{tikzpicture}

		{\footnotesize$\Delta \bm{p}_t$ is the distance of the object to the goal position, $\Delta h$ is the clearance between the object and the table, $\bm{c}_t$ is the vector of contact forces acting on the robot arm, and $r_v$ is the visibility ratio of the target object.
					}
      \vspace{-0.3cm}	
\end{table}

\section{Task and System Description}
We study the problem of learning to grasp objects with an anthropomorphic robot hand in both the isolated single-object case and from more cluttered environments.
Figure~\ref{fig:robot_setup} shows the robot setup, which we replicate in Isaac Gym~\cite{Makoviychuk2021}.

\subsection{Task Formulation}
\subsubsection{Observation and action space}
The teacher policy takes as input proprioceptive information from the robot, the ground truth state of the object, and its own previous action.
For the student policy, the ground truth state of the object is replaced by a segmented point cloud of the target object.
Actions $\bm{a}_t \in \mathbb{R}^{11}$ specify desired changes to the end-effector's 6D pose and the joint angles of the fingers. 
The control frequency of the system is 7.5 Hz and an episode terminates after 75 steps.

\subsubsection{Reward function and success criterion}
The reward function combines incentives for task completion with auxiliary objectives to guide exploration and maintain safety.
Exploration is facilitated by rewarding the policy for bringing the fingertips close to the target object and for lifting the object off the table. To learn safe behaviors, we penalize large actions and contact forces acting on the robot arm.
The full reward function is detailed in Table~\ref{tab:reward_terms}.

We consider an episode as successful if the target object is lifted to within 5\,cm of its target position central above the table.
The initial poses of the objects are randomized by dropping them sequentially until all objects are in the workspace of the robot.

\subsubsection{Evaluated objects} We utilize the YCB object and model set~\cite{Calli2015}, which contains diverse every-day items, to train our policies.
Some YCB objects are too large to be grasped with our Schunk SIH hand. 
Therefore, we select a subset of graspable objects with at least two bounding-box dimensions measuring under 10\,cm.
This results in a total of 60 objects, which we randomly divide into 48 training objects and 12 test objects.

\begin{table}[tb]
    \centering 
    \caption{Success rates (\%) of the teacher policies.}
    {\small \begin{tabular}{l|ccc}
            \toprule 
            & \multicolumn{3}{c}{Number of objects} \\
            Object set & 1 & 3 & 5 \\
            \midrule
            train &  $95.1 \pm 0.5$ & $85.3 \pm 0.7$ & $79.0 \pm 1.1$ \\
            test &  $89.7 \pm 0.7$ & $82.2 \pm 0.6$ & $75.9 \pm 1.4$ \\

            \bottomrule
        \end{tabular}
		}		\vspace*{-2ex}
    \label{tbl:teacher_policy_performance}
\end{table}

\subsection{System Description}

\subsubsection{System setup}
Our setup employs a Universal Robots UR5 robotic arm paired with a Schunk SIH five-finger hand. The hand features 11 degrees of freedom, of which 5 are tendon-actuated.
For visual perception, we use a RealSense D455 RGB-D camera mounted statically above the workspace.

\subsubsection{Policy architecture}
The teacher policy consists of an MLP with three hidden layers, comprising 1024, 512, and 512 units, respectively. 
Each layer uses ELU activations~\cite{Clevert2015}.
The student policy architecture splits the processing of vectorized and point cloud-based observations to obey the permutation invariant nature of this observation type.
We employ a PointNet-like encoder~\cite{Qi2017} to process the point cloud and concatenate the resulting embedding with the remaining observations.
The result is fed into an MLP with the same architecture as the teacher policy.

\begin{figure}[t]
    \centering
        \includegraphics[trim={0 2.5mm 0 0},clip, width=\linewidth]{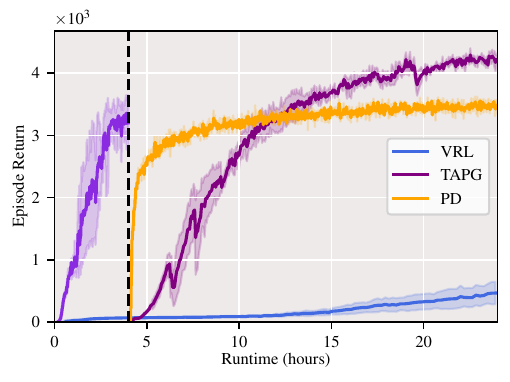}
				\vspace*{-3ex}
        \label{fig:learning_curves_single_object_grasping}
    \caption{Episode return during training illustrating learning speed and final performance of different methods for obtaining a sensorimotor grasping policy.
    Vertical line indicates transition from privileged teacher (in light purple) to student policy.
    We limit the maximum training time to 24 hours and report the mean and standard deviation of three runs.\vspace*{-2ex}
    }
    \label{fig:learning_curves}
\end{figure}

\section{Results}

Our experimental design is set up to evaluate the efficacy of our approach in dexterous robotic grasping. 
Specifically, we target the following research questions: 
(1) Is the teacher policy able to master dexterous grasping using privileged state information?
~(2) Can TAPG distill the teacher's expertise into a sensorimotor student policy and demonstrate a tangible advantage over the baselines?
~(3) Are the resultant policies capable of real-world deployment?

The rationale of our proposed method naturally suggests a comparison with two methods for acquiring a sensorimotor policy.
The first is to apply RL directly from visual observations, which we denote as VRL. 
The second, labeled PD, leverages policy distillation. 
However, unlike our method, it solely relies on cloning the teacher's behavior, without any further fine-tuning of the sensorimotor behaviors.

\subsection{Training the Teacher Policy}
We train our teacher policies on YCB objects selected uniformly from the training set.
As reported in Table~\ref{tbl:teacher_policy_performance}, the resultant policies achieve a success rate of over 95\% for the single-object case.
Moreover, our evaluations of zero-shot transfer to the test set underline the robustness of the learned behaviors.
We further probe their robustness by placing 3 or 5 items in the workspace to induce unseen interactions, resulting in success rates of 85.3\% and 79.0\%, respectively, when using training objects and 82.2\% and 75.9\% when using test objects.
A qualitative investigation of failure modes showed that the majority of unsuccessful trials result from objects that are very flat, making it difficult to pick them up.

Overall, the privileged teacher policies adeptly solve all task configurations, answering our first research question in the affirmative. 
We can confirm that model-free RL is able to synthesize proficient control policies for the investigated problem given access to privileged simulator state information and ample environment interactions.

\begin{figure}[t]
    \textit{Policy distillation (PD)}
    \vspace{0.075cm}
    \setlength{\fboxsep}{0pt}

    \begin{subfigure}{0.245\linewidth}
        \centering
        \colorbox{grey_background}{\includegraphics[trim={0 1cm 0 0},clip, width=\linewidth]{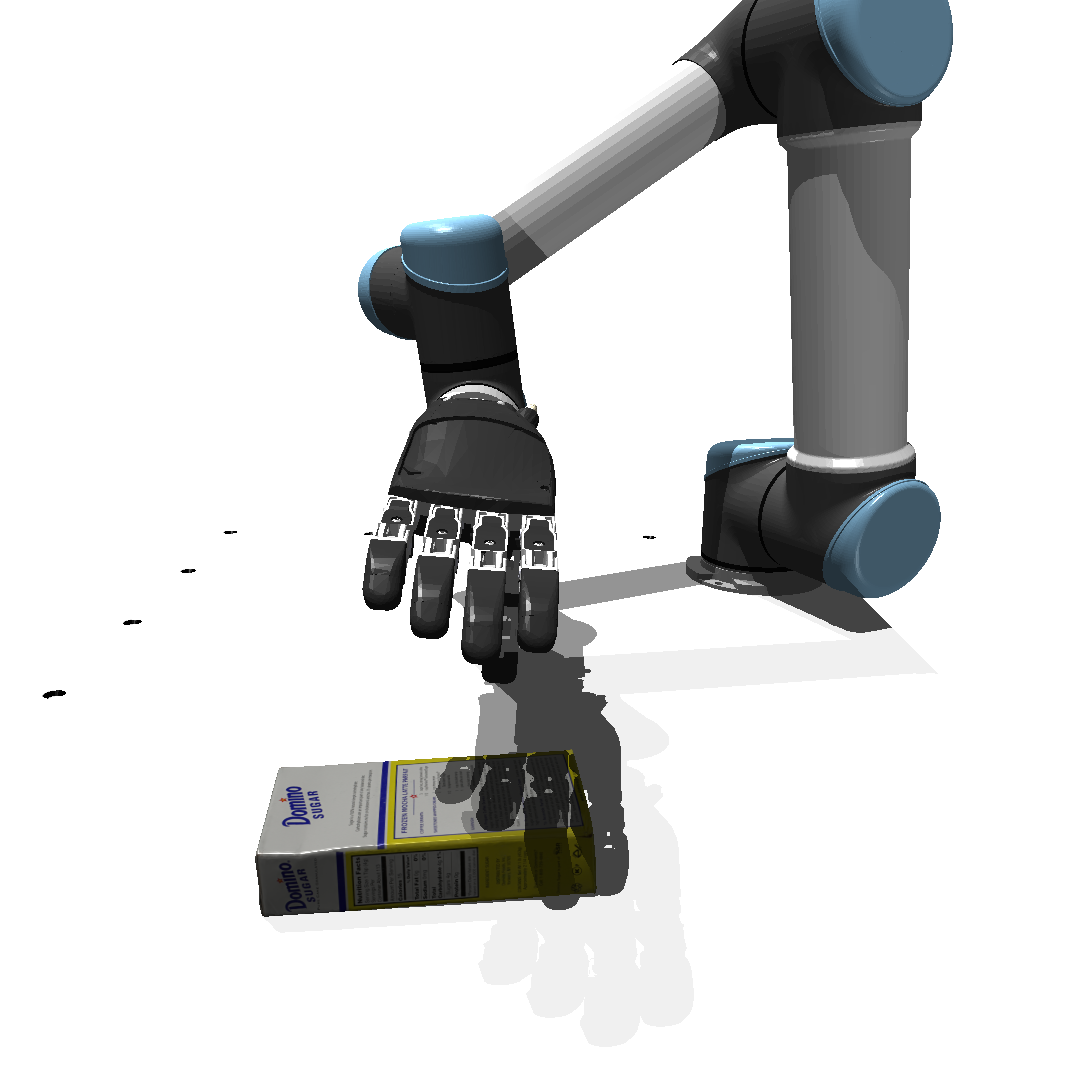}}
    \end{subfigure}%
    \hfill
    \begin{subfigure}{0.245\linewidth}
        \centering
        \colorbox{grey_background}{\includegraphics[trim={0 1cm 0 0},clip, width=\linewidth]{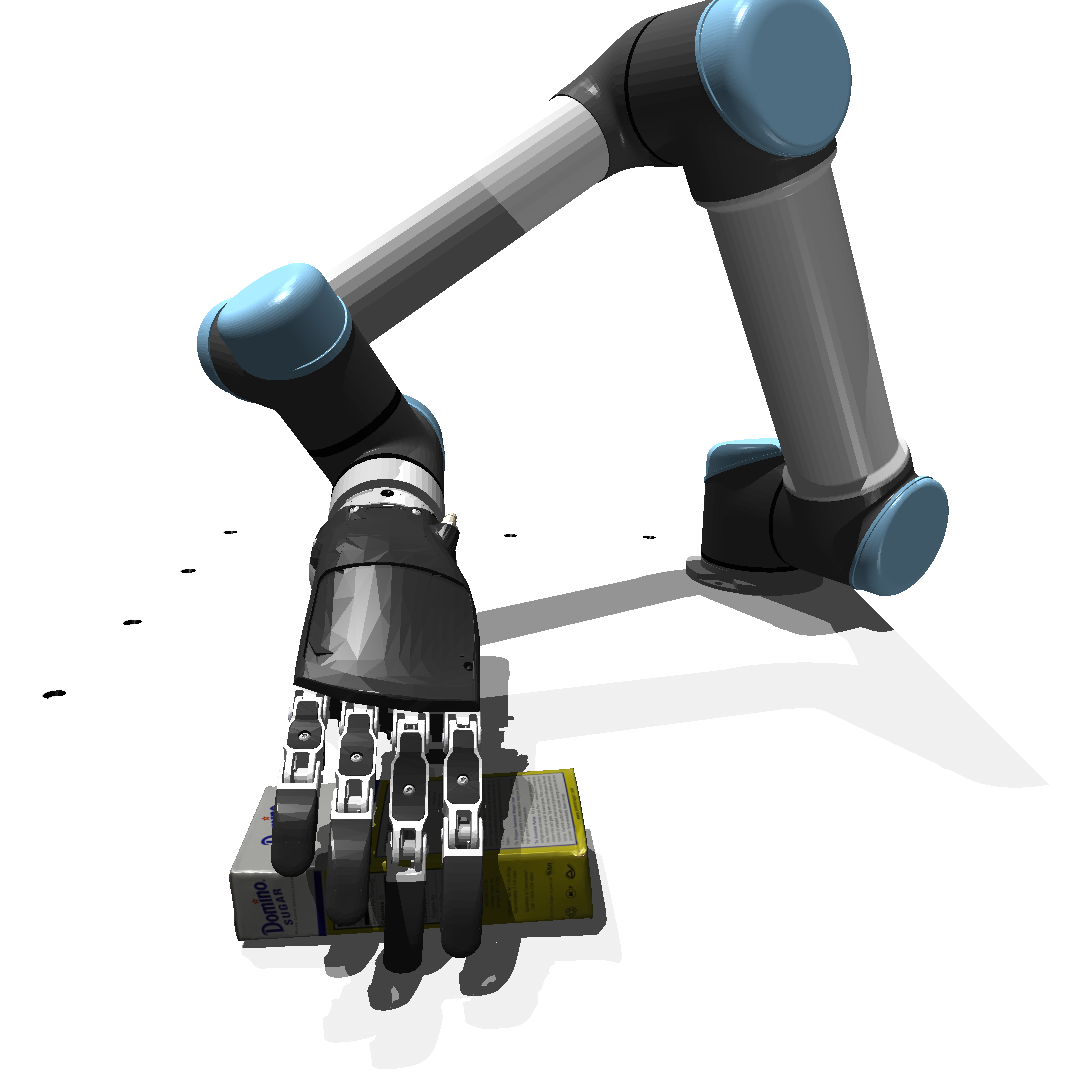}}
        \end{subfigure}%
    \hfill
    \begin{subfigure}{0.245\linewidth}
        \centering
        \colorbox{grey_background}{\includegraphics[trim={0 1cm 0 0},clip, width=\linewidth]{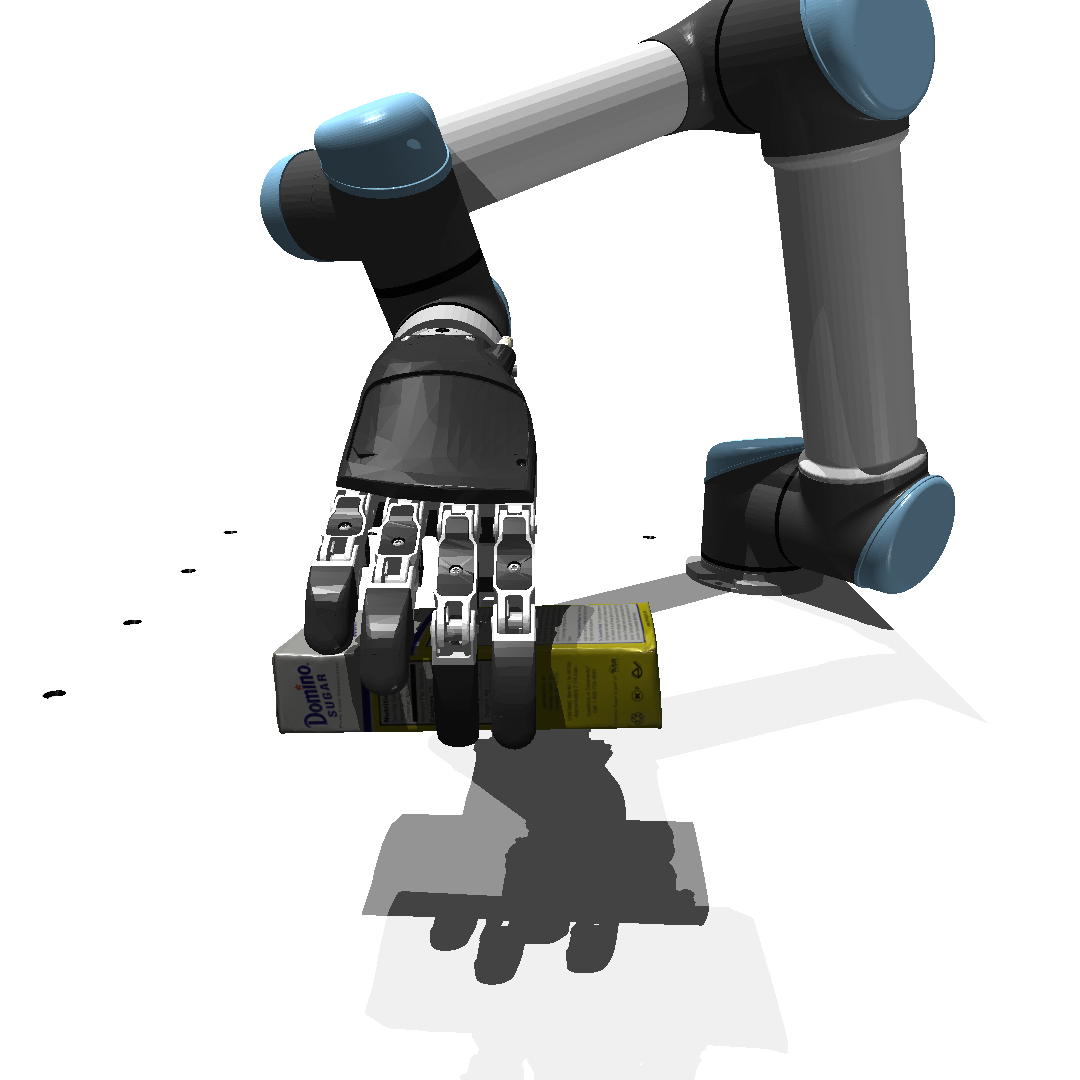}}
        \end{subfigure}%
    \hfill
    \begin{subfigure}{0.245\linewidth}
        \centering
        \colorbox{grey_background}{\includegraphics[trim={0 1cm 0 0},clip, width=\linewidth]{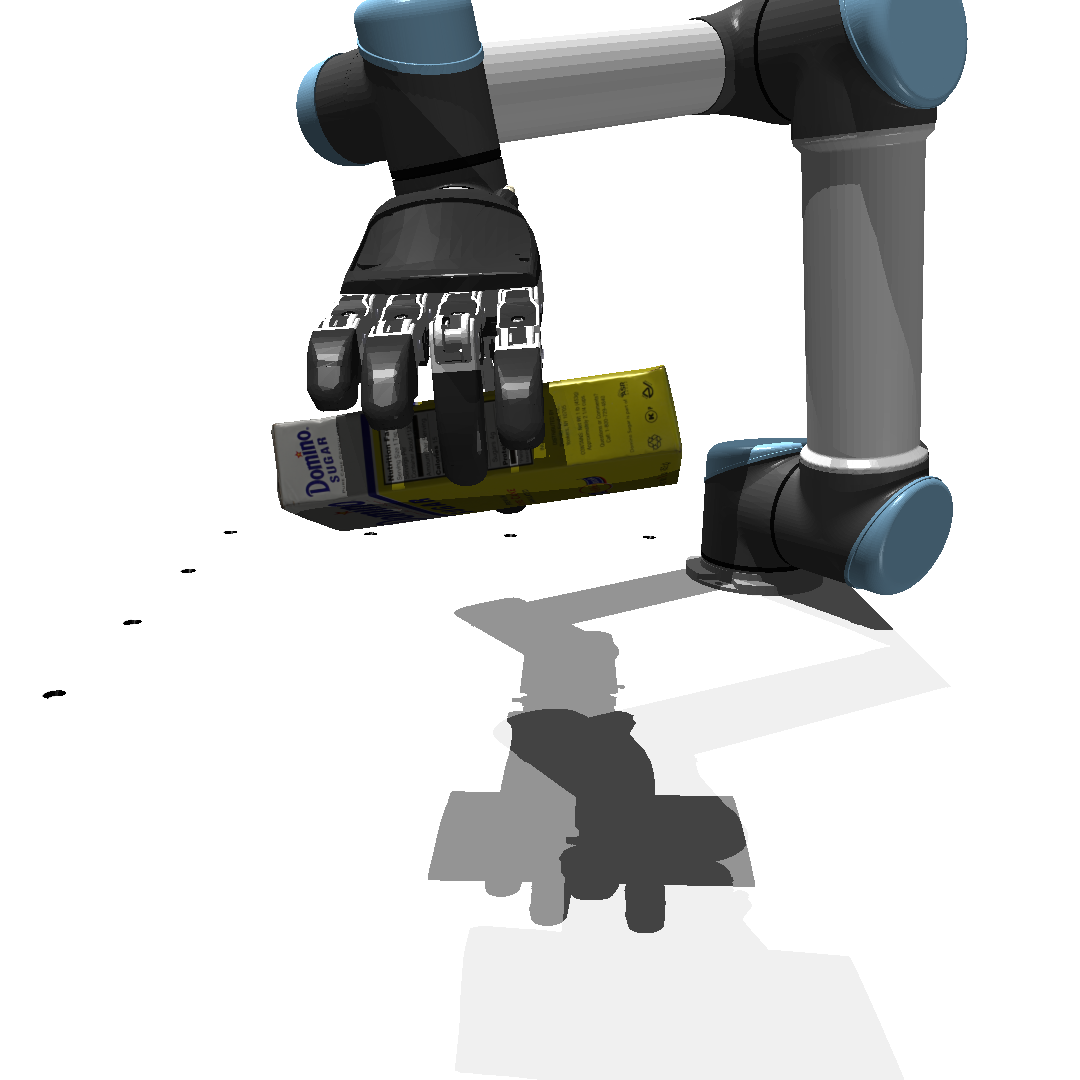}}
    \end{subfigure}%

    \begin{subfigure}{0.245\linewidth}
            \centering
            \colorbox{rose_background}{\includegraphics[trim={0 3cm 0 0},clip, width=\linewidth]{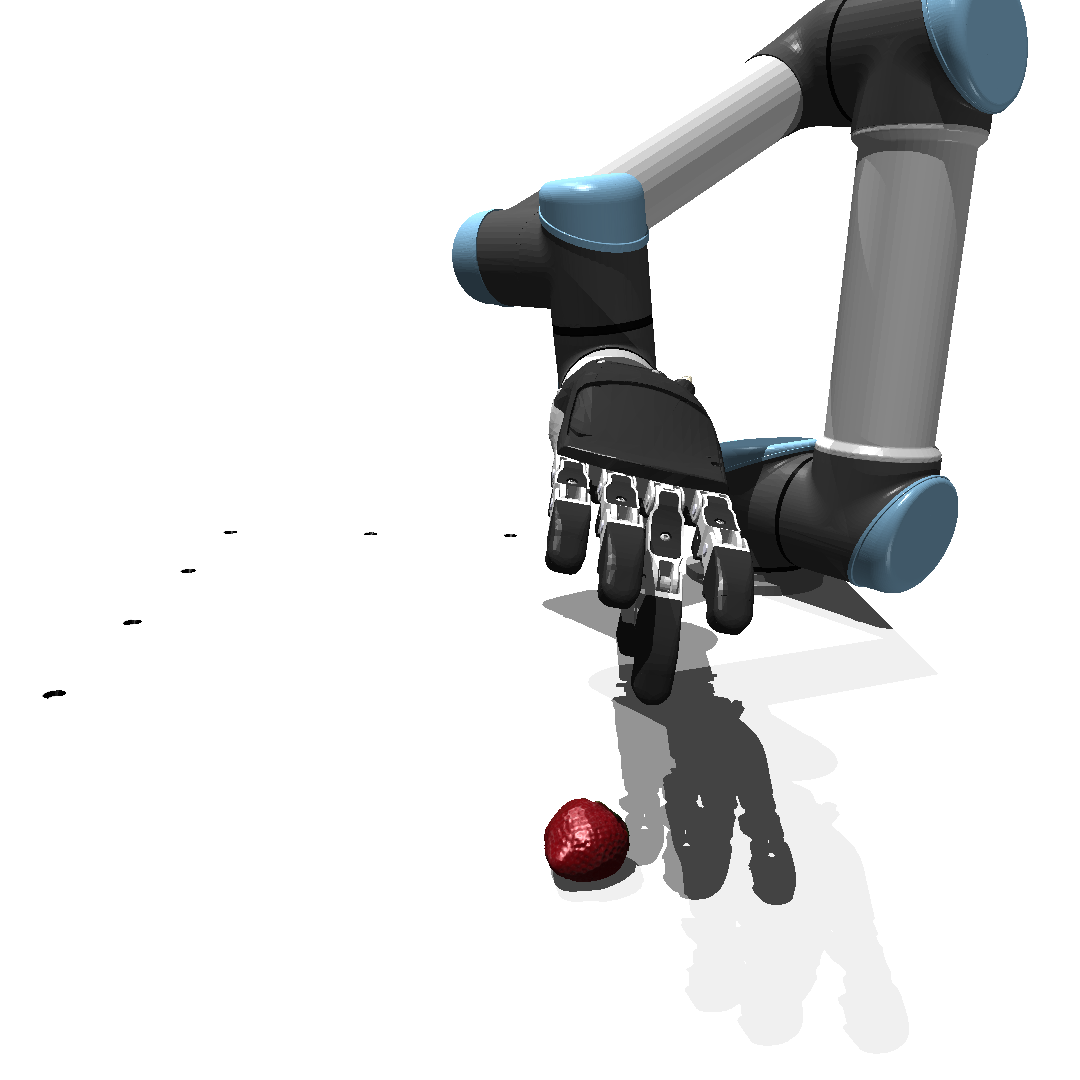}}
        \end{subfigure}%
        \hfill
        \begin{subfigure}{0.245\linewidth}
            \centering
            \colorbox{rose_background}{\includegraphics[trim={0 3cm 0 0},clip, width=\linewidth]{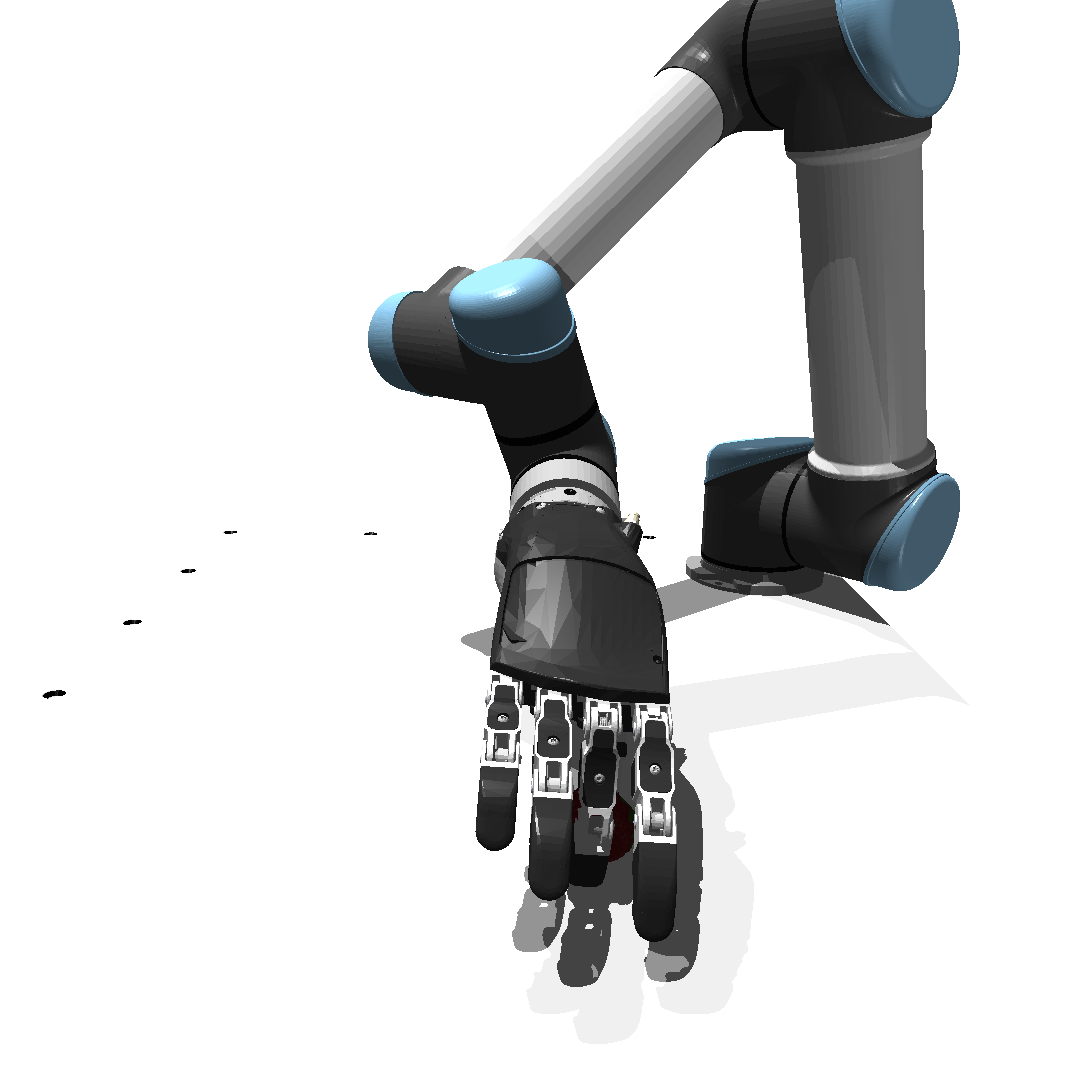}}
            \end{subfigure}%
        \hfill
        \begin{subfigure}{0.245\linewidth}
            \centering
            \colorbox{rose_background}{\includegraphics[trim={0 3cm 0 0},clip, width=\linewidth]{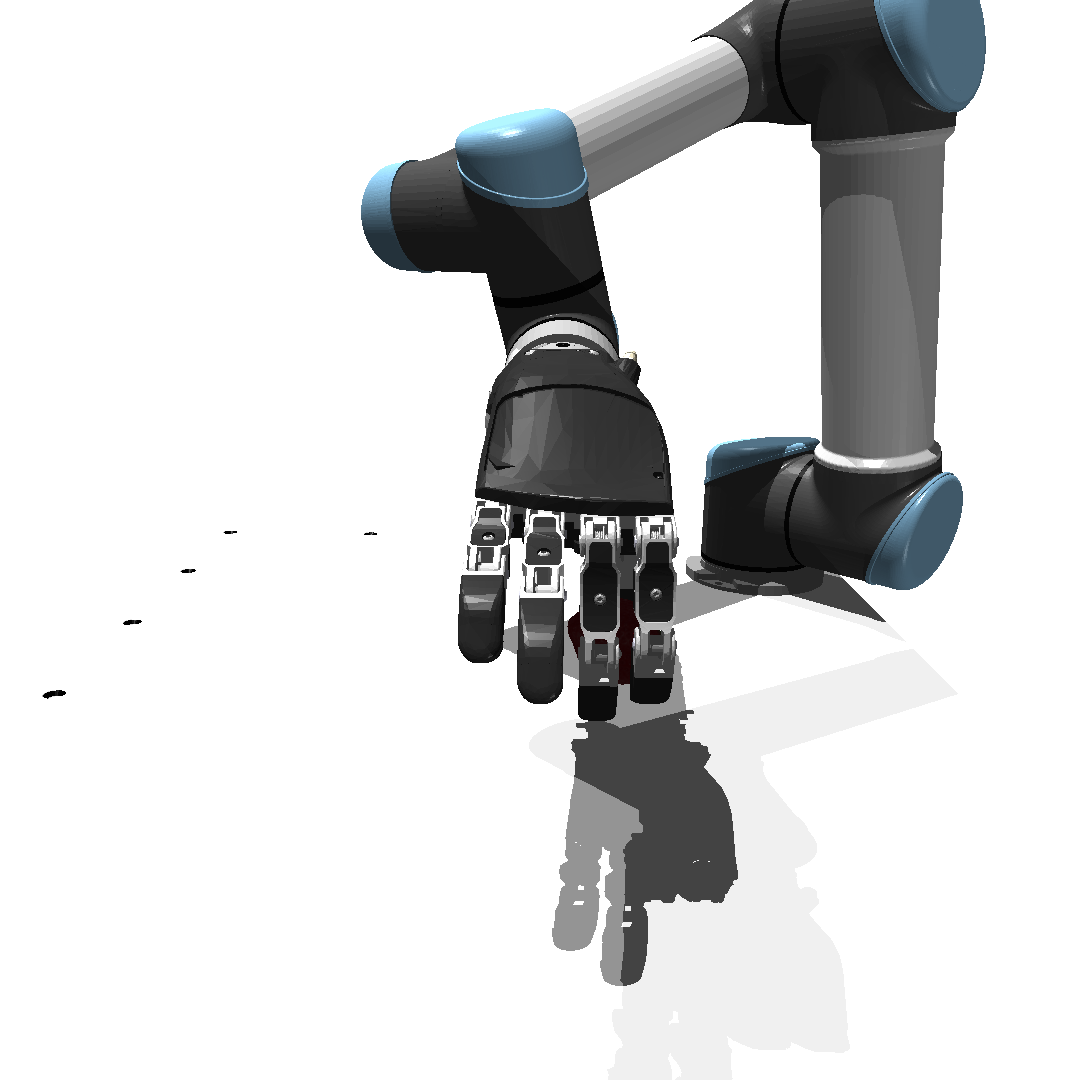}}
            \end{subfigure}%
        \hfill
        \begin{subfigure}{0.245\linewidth}
            \centering
            \colorbox{rose_background}{\includegraphics[trim={0 3cm 0 0},clip, width=\linewidth]{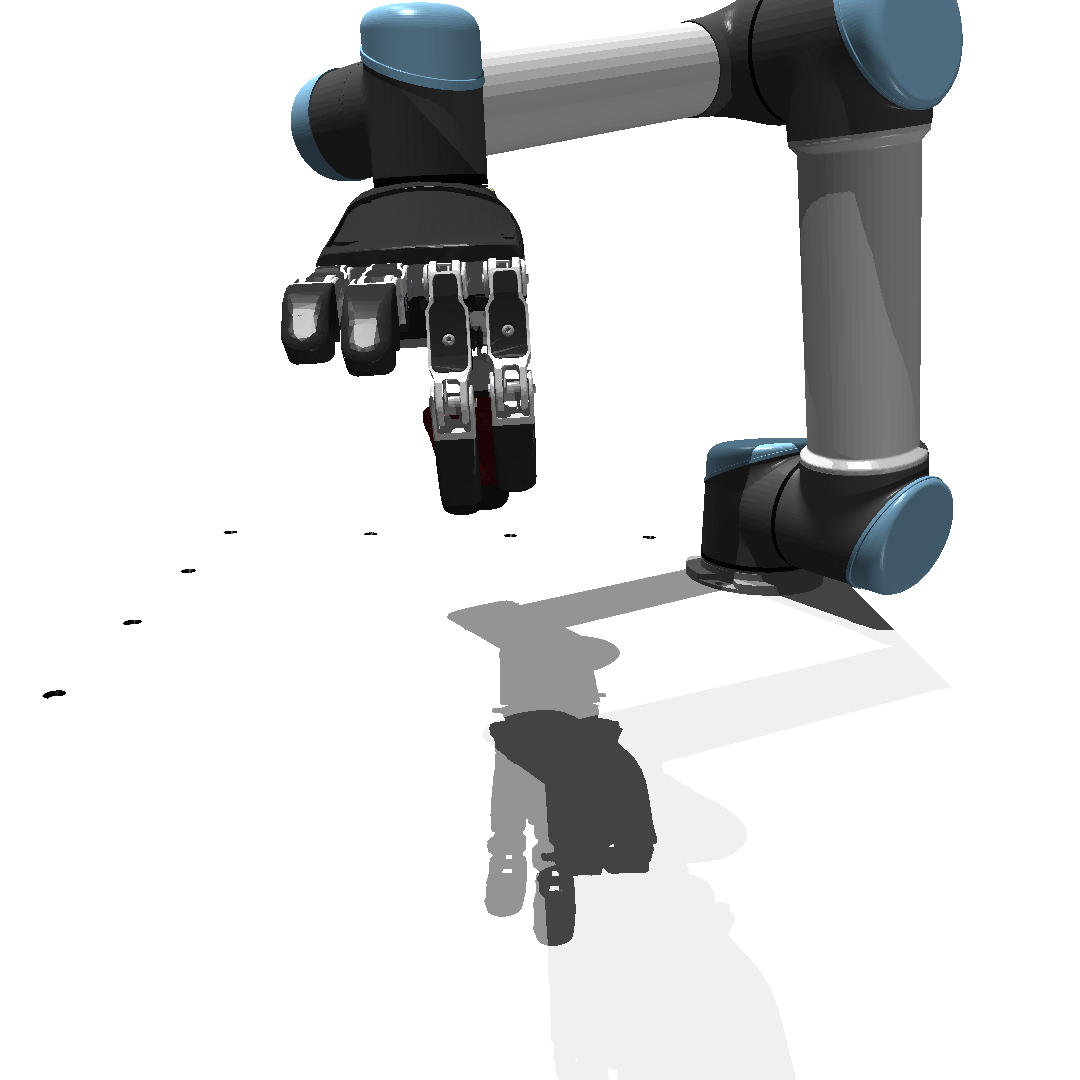}}
        \end{subfigure}%

    \textit{Teacher-Augmented Policy Gradient (TAPG, ours)}
    \vspace{0.075cm}

    \begin{subfigure}{0.245\linewidth}
            \centering
            \colorbox{grey_background}{\includegraphics[width=\linewidth]{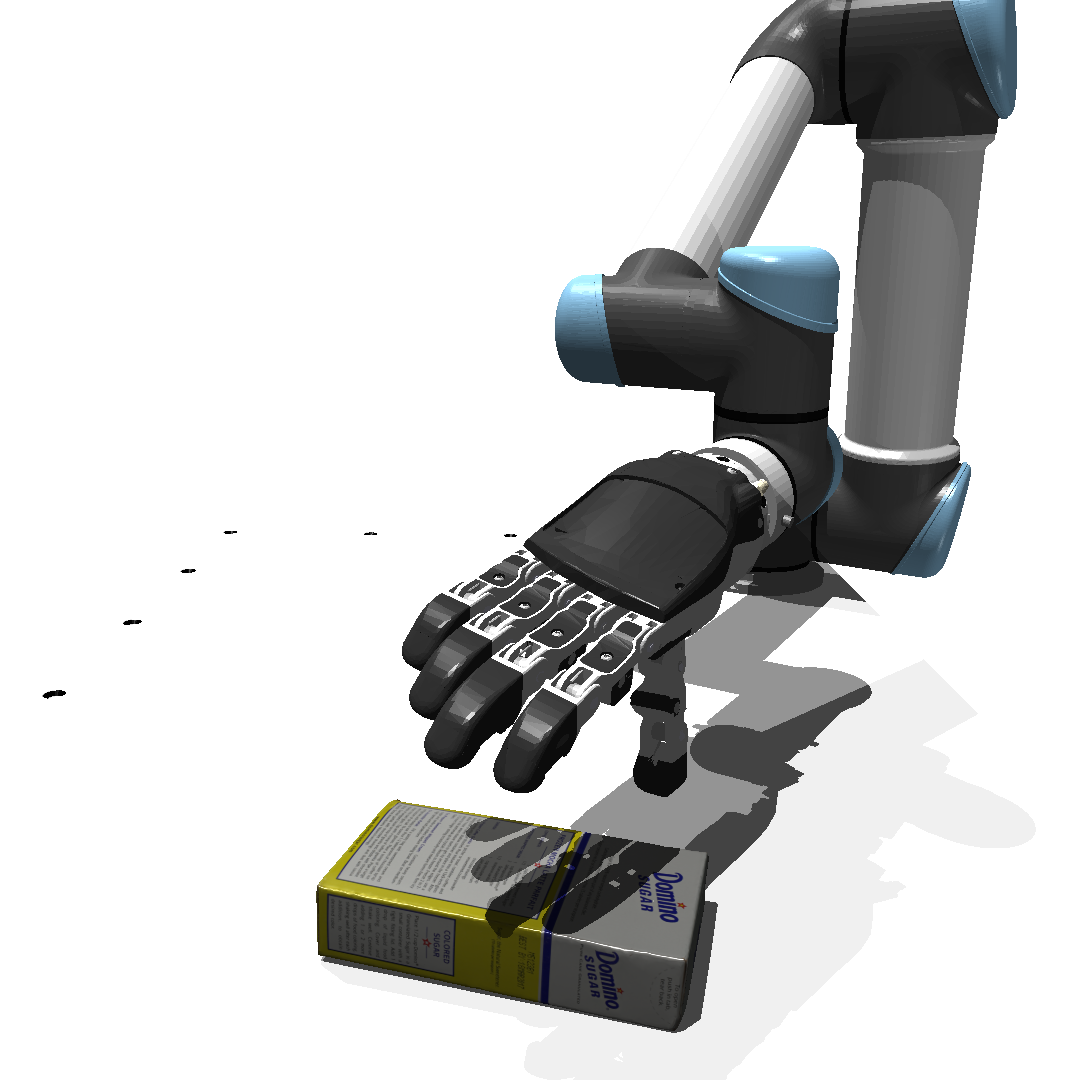}}
        \end{subfigure}%
        \hfill
        \begin{subfigure}{0.245\linewidth}
            \centering
            \colorbox{grey_background}{\includegraphics[width=\linewidth]{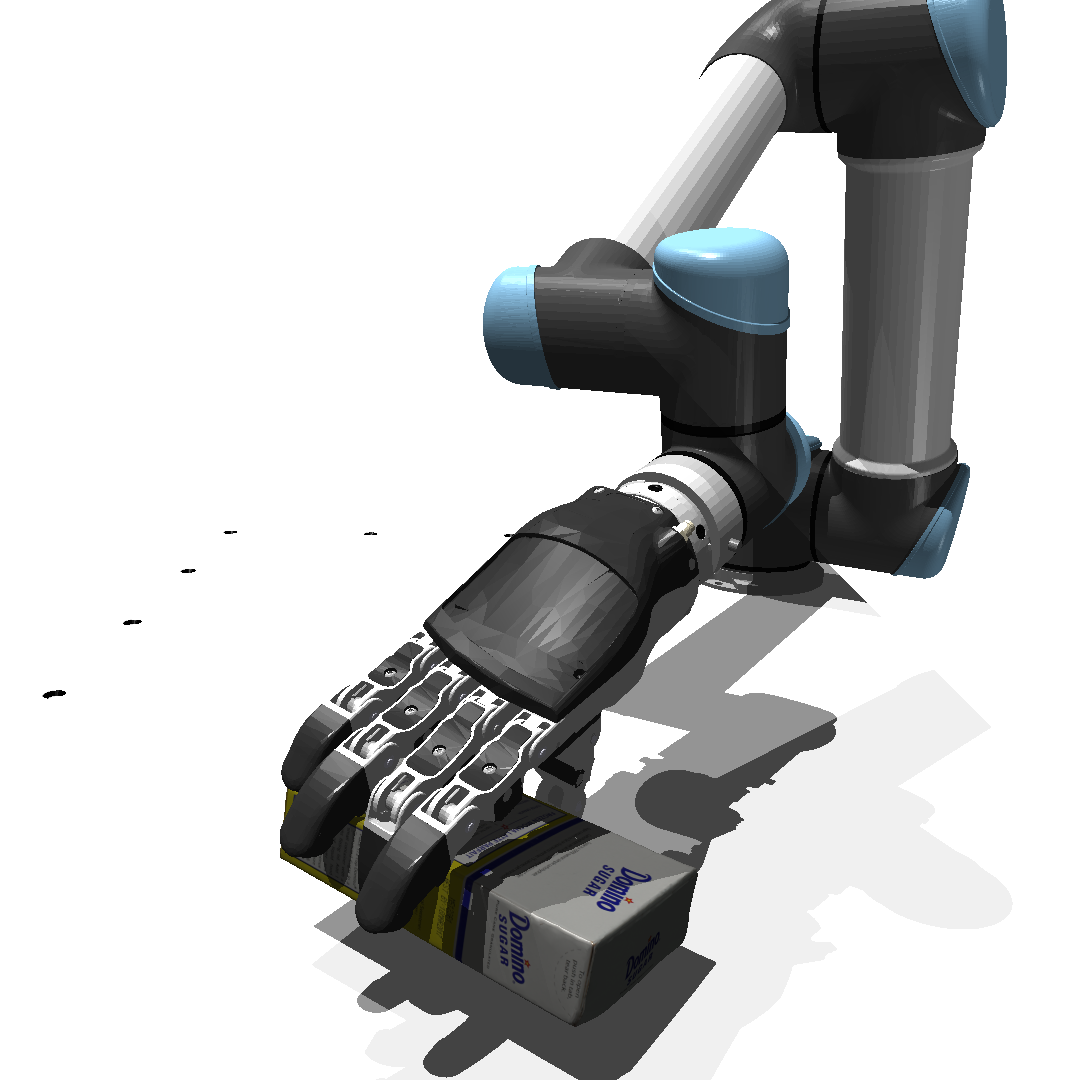}}
            \end{subfigure}%
        \hfill
        \begin{subfigure}{0.245\linewidth}
            \centering
            \colorbox{grey_background}{\includegraphics[width=\linewidth]{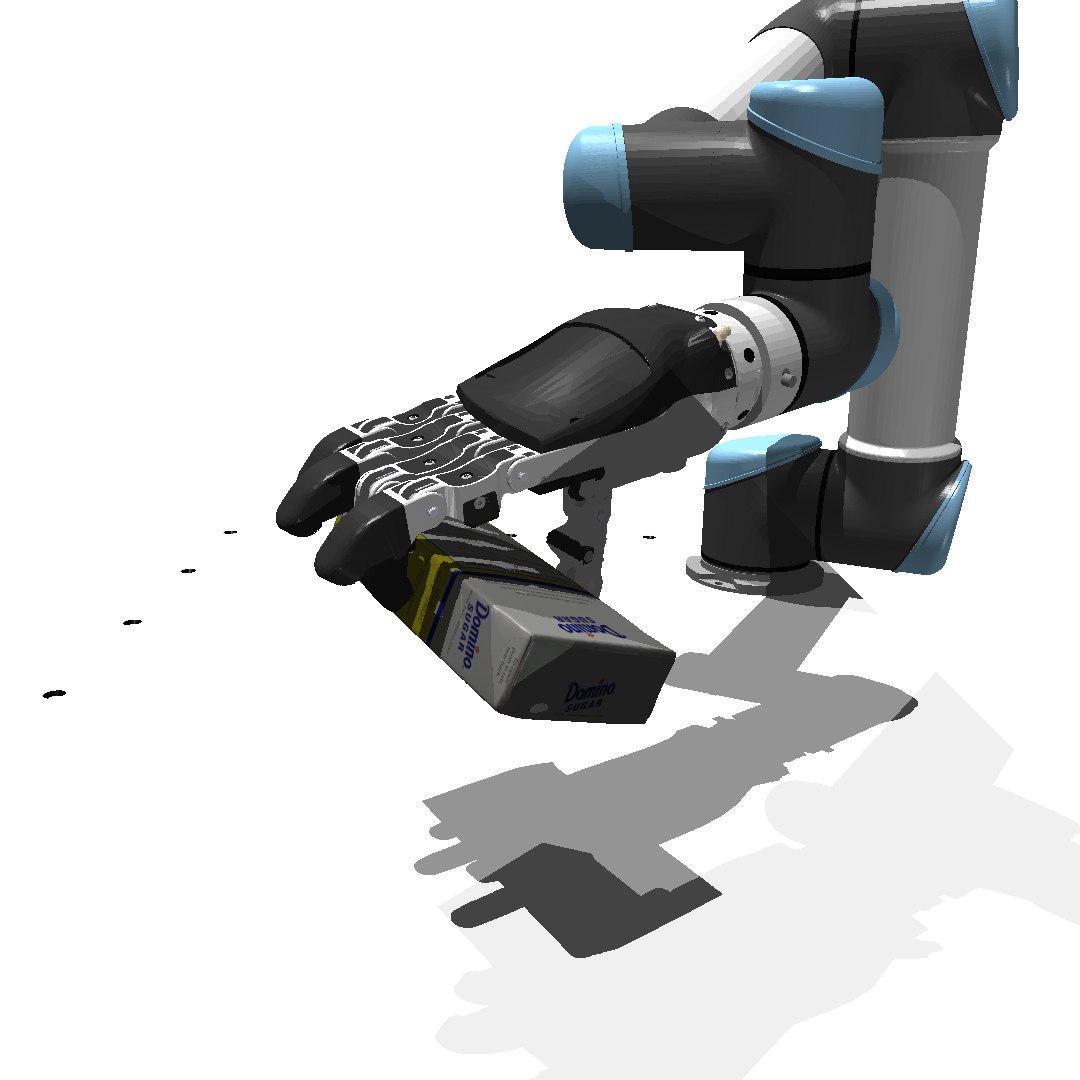}}
            \end{subfigure}%
        \hfill
        \begin{subfigure}{0.245\linewidth}
            \centering
            \colorbox{grey_background}{\includegraphics[width=\linewidth]{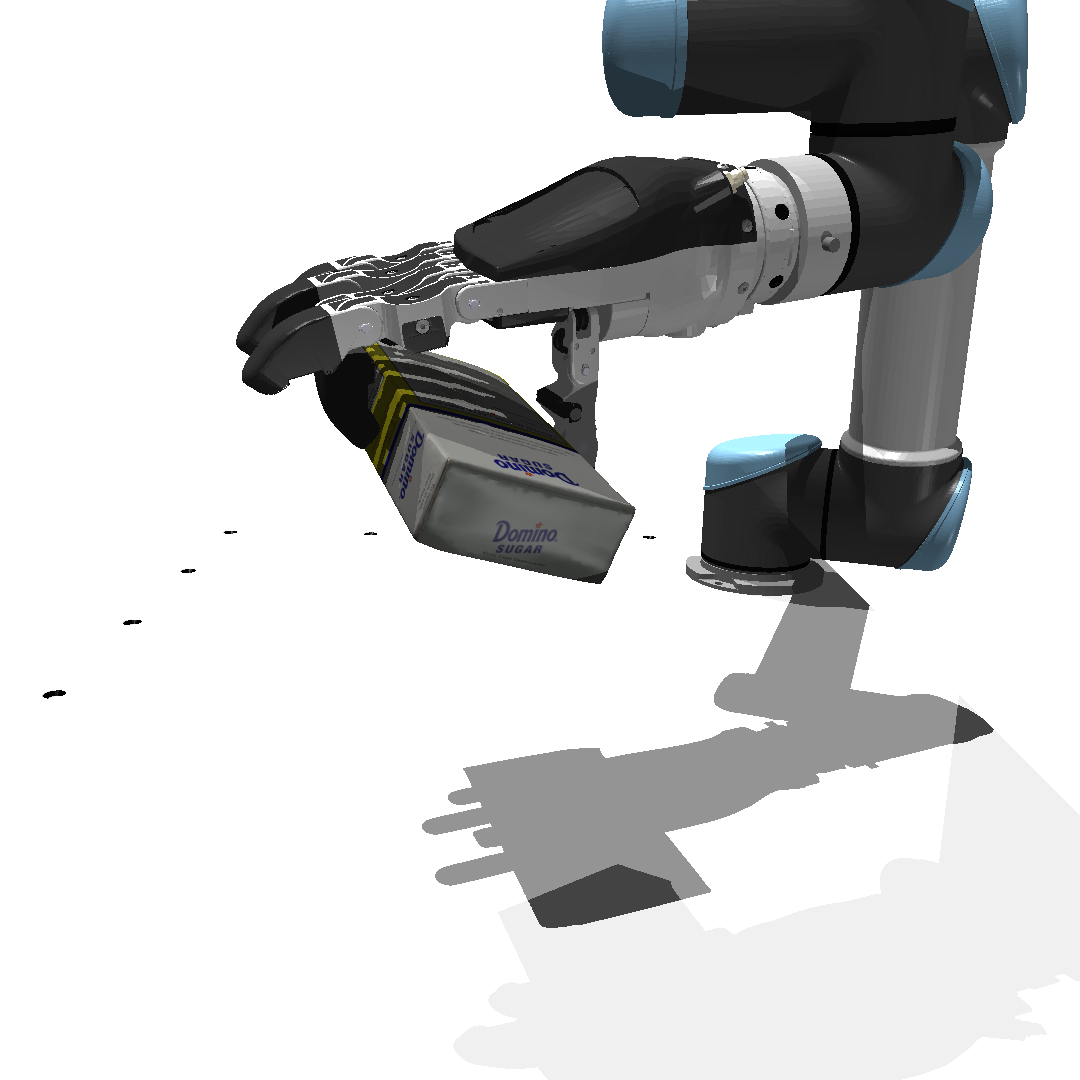}}
        \end{subfigure}%

        \begin{subfigure}{0.245\linewidth}
                \centering
                \colorbox{rose_background}{\includegraphics[trim={0 5cm 0 0},clip, width=\linewidth]{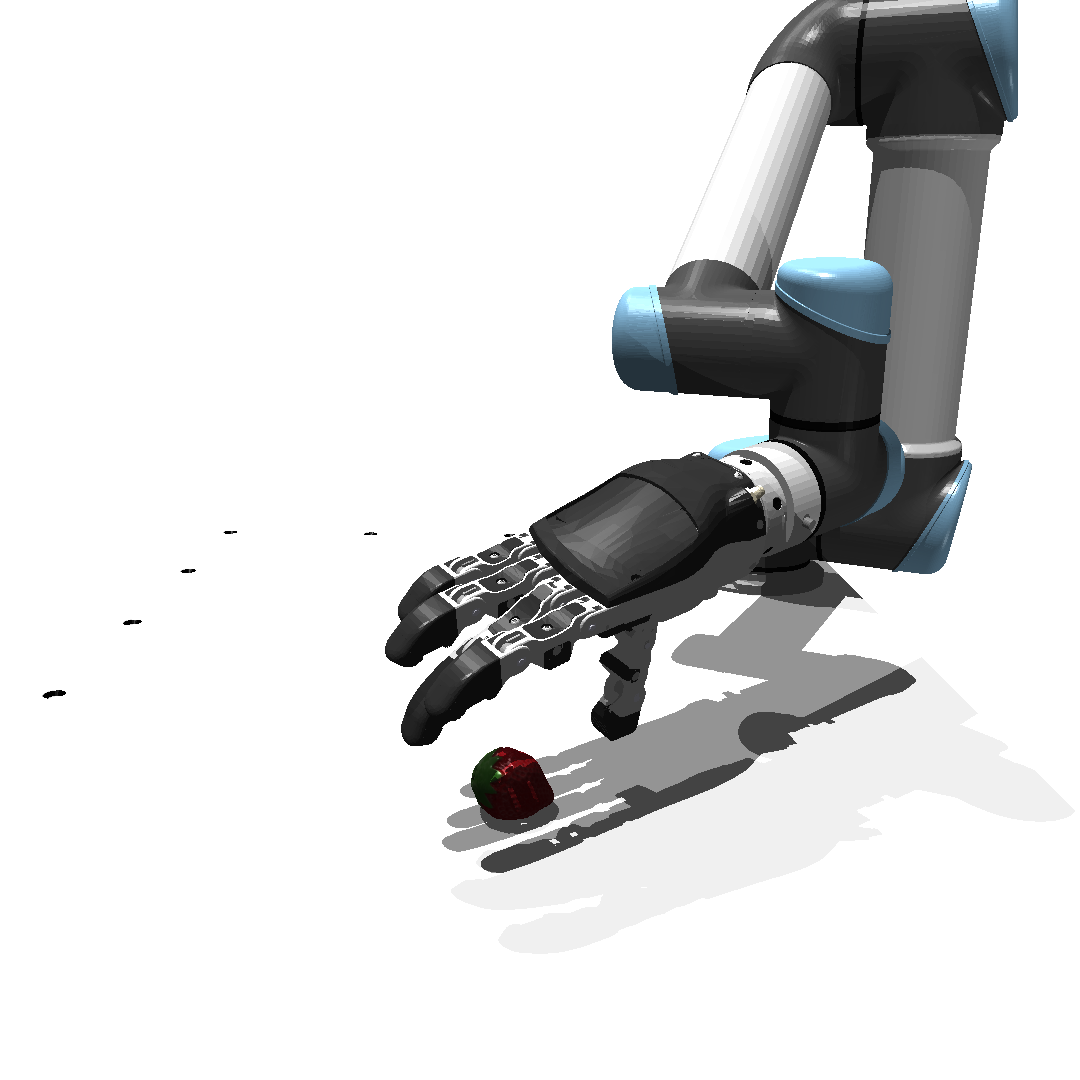}}
            \end{subfigure}%
            \hfill
            \begin{subfigure}{0.245\linewidth}
                \centering
                \colorbox{rose_background}{\includegraphics[trim={0 5cm 0 0},clip, width=\linewidth]{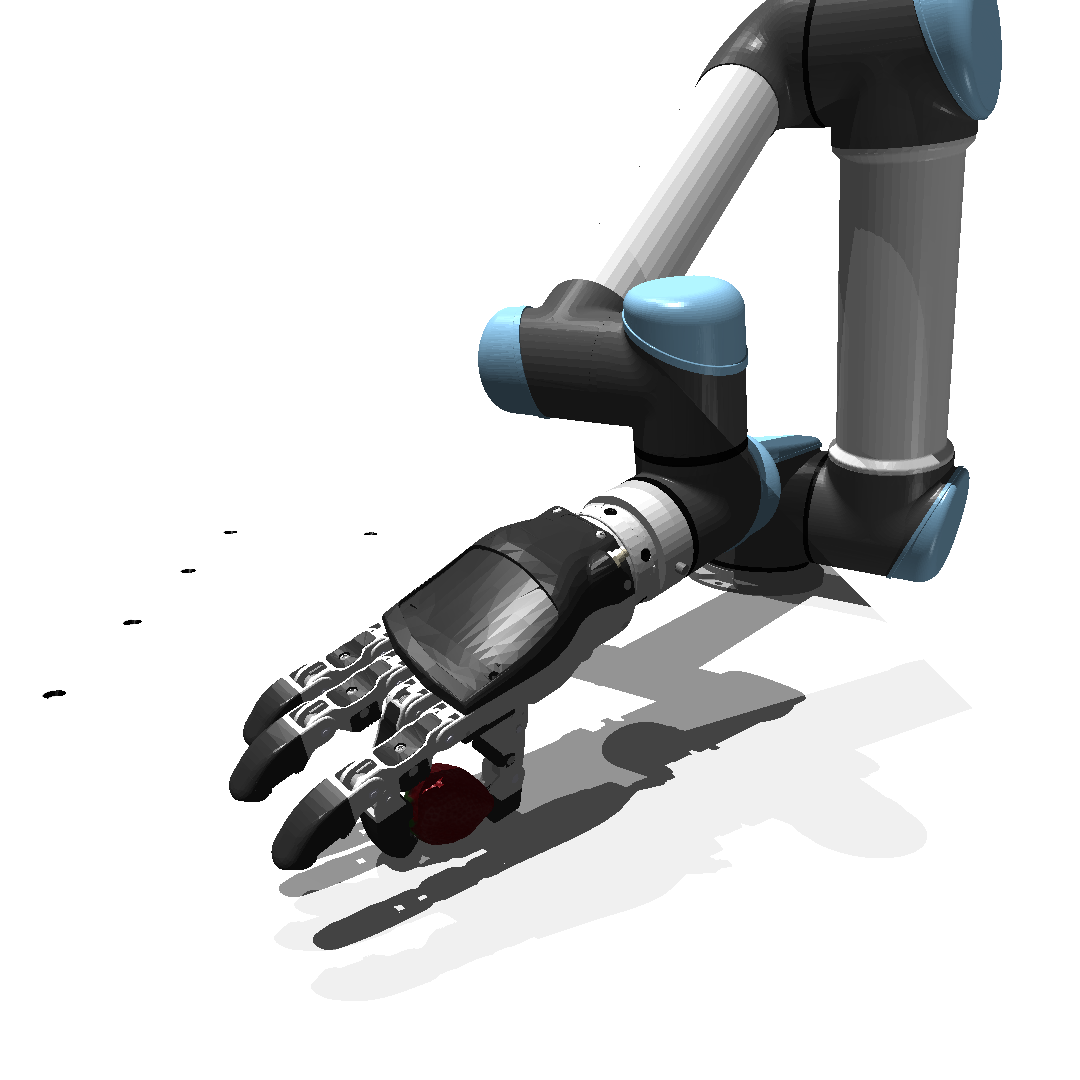}}
                \end{subfigure}%
            \hfill
            \begin{subfigure}{0.245\linewidth}
                \centering
                \colorbox{rose_background}{\includegraphics[trim={0 5cm 0 0},clip, width=\linewidth]{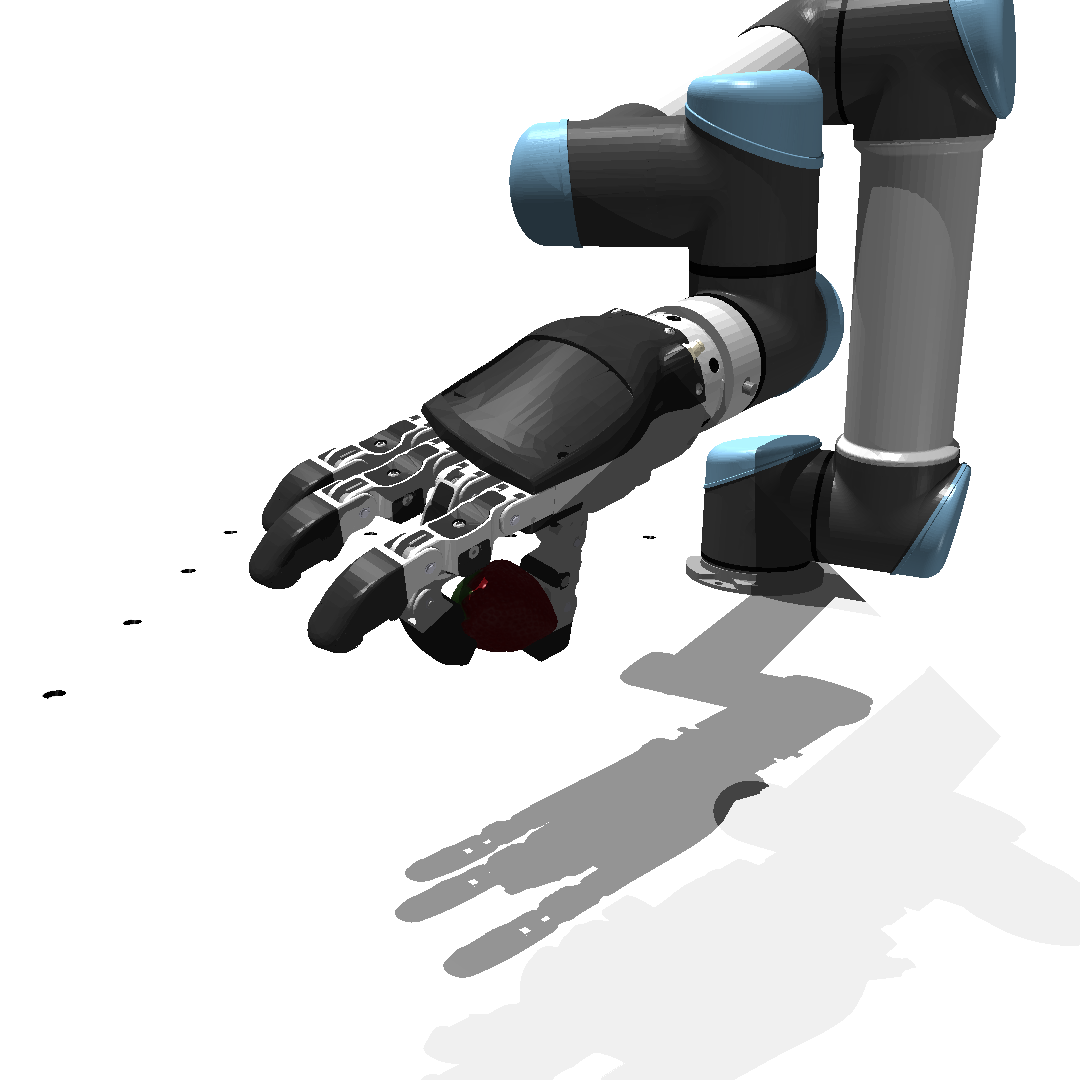}}
                \end{subfigure}%
            \hfill
            \begin{subfigure}{0.245\linewidth}
                \centering
                \colorbox{rose_background}{\includegraphics[trim={0 5cm 0 0},clip, width=\linewidth]{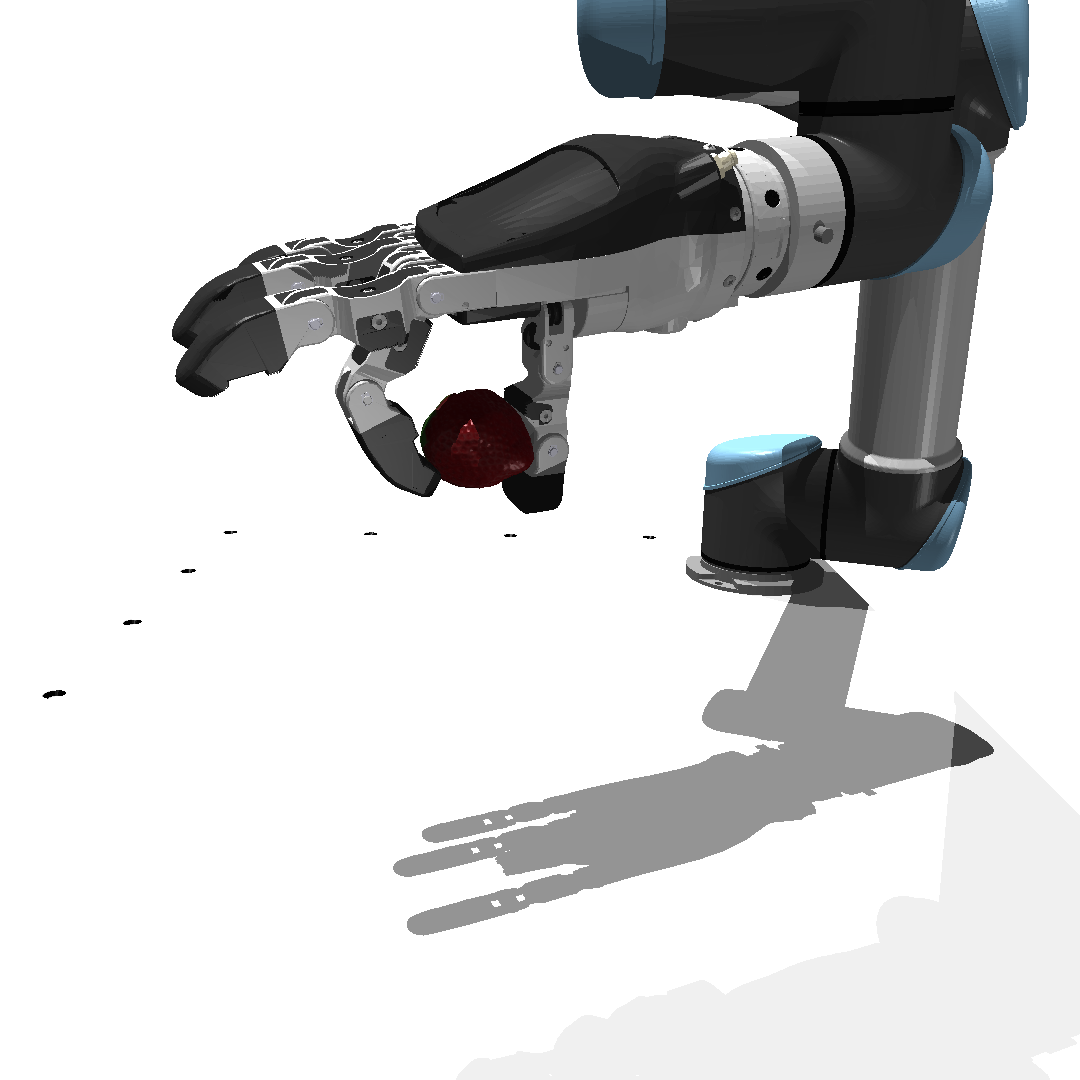}}
            \end{subfigure}%
            
    \caption{Grasping behaviors learned by vanilla PD versus TAPG as viewed from the camera used for tracking.
    We see that TAPG assumes the learned grasping skills and successfully adapts them to the visual observation space.
    For larger objects, such as the sugar box (top rows), this amounts to retracting the fingers not actively involved in holding the object and grasping at an angle. 
    Here, both methods learn behaviors that retain the visibility of the majority of the object.
    In contrast, for smaller objects such as the strawberry (bottom rows), modifying the behaviors becomes crucial.
    While the PD policy severely occludes the object, TAPG maintains object visibility throughout the grasping process.\vspace*{-3ex}
    }
    \label{fig:pd_vs_tapg_learned_grasps}
\end{figure}

\subsection{Training the Sensorimotor Policy}
In this section, we analyze different methods for obtaining a sensorimotor policy.
Specifically, we contrast the performance of two established frameworks (VRL and PD) with our proposed method TAPG.
The progression of the policy performance is plotted in Figure~\ref{fig:learning_curves}.
RL from vision (VRL), while making some progress, cannot solve the studied task in any acceptable time.
Both PD and TAPG learn capable sensorimotor policies. 
The fully supervised learning paradigm of PD results in fast and stable convergence of the student policy.
Although TAPG requires longer to recover the teacher's performance, it is subsequently able to surpass it.
As such, TAPG is the only paradigm able to sufficiently adapt to the intricacies and requirements of visual perception.

To elucidate the improvement of TAPG over PD, we depict the grasping behaviors learned by both paradigms in Figure~\ref{fig:pd_vs_tapg_learned_grasps}.
TAPG is able to improve the visibility of the target object while maintaining high grasping performance.
This effect is most pronounced for small objects that fit inside the hand.
In summary, our findings show that TAPG offers substantial benefits over the baselines for the studied task, thereby affirmatively answering our second question.

\begin{table}[tb]
    \centering
    \caption{Quantitative zero-shot real-world evaluation.}
 {\small   \begin{tabular}{l|cccc}
        \toprule
        & \multicolumn{4}{c}{Number of objects} \\
        Method & 1 & 3 & 5 & ALL \\
        \midrule
        Visual RL (VRL) &  \large \;\sfrac{0}{60} & \large \;\sfrac{0}{10} & \large \;\sfrac{0}{10} & \large \;\sfrac{0}{80} \\
        Policy distillation (PD) &  \large \;\sfrac{18}{60} & \large \;\sfrac{2}{10} & \large \;\sfrac{0}{10} & \large \;\sfrac{20}{80} \\
        TAPG (ours) & \textbf{\large \;\sfrac{35}{60}} & \textbf{\large \;\sfrac{5}{10}} & \textbf{\large \;\sfrac{4}{10}} & \textbf{\large \;\sfrac{44}{80}} \\
        \bottomrule
    \end{tabular}				
	}	\\ \vspace*{1ex} {\footnotesize Reported are ~\raisebox{.1mm}{successful\,trials}\,$\sfrac{}{}$\,\raisebox{-.1mm}{number\,of\,trials}.}
    \label{tbl:real_world_performance} 
\end{table}

\subsection{Real-world Deployment}
The proposed pipeline transfers well to the real world, without any adaptation (zero-shot).
After manually placing one or multiple objects in the robot's workspace, we deem a trial successful if the robot retrieves and holds the target object above the table.
We perform five runs for each test object and ten runs for cluttered configurations with three and five objects on the table.
The results are reported in Table~\ref{tbl:real_world_performance}. While VRL is not successful, PD succeeds in 25\% of the trials. 
Our method TAPG clearly outperforms PD by succeeding in 55\% of the trials.

Next, to verify the benefit of the adapted behaviors, we report an object-wise performance assessment in Table~\ref{tbl:object_wise}.
For large objects that are never fully occluded (bleach cleanser and sugar box), maintaining the behaviors of the teacher policy proves to be sufficient.
In contrast, the strength of TAPG can really be seen on the smaller objects (foam brick and smallest cup) that can fit inside the hand.
Occluding them fully during the grasping process causes the vision pipeline to lose track of the object, resulting in failed executions.

\begin{table}[tb]
        \centering
        \caption{Object-wise zero-shot real-world evaluation.}
 {\small       \begin{tabular}{M{0.75cm}lcc}
            \toprule
            & Object & PD & TAPG (ours) \\
            \midrule
            \includegraphics[trim={0 3cm 0 3cm},clip,height=1.5em]{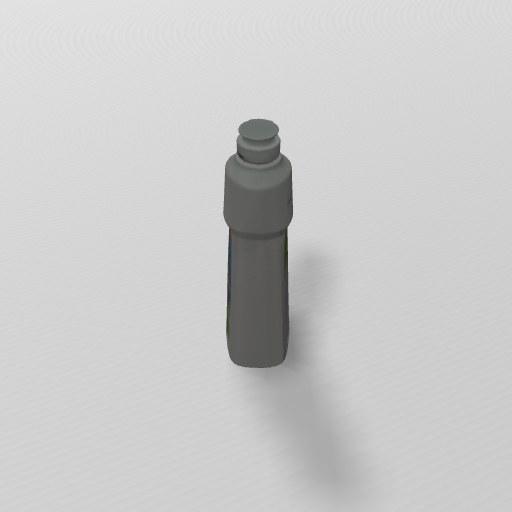} & 021\_bleach\_cleanser & \large \;\sfrac{4}{5} & \textbf{\large \;\sfrac{5}{5}} \\
            \vspace{-0.1cm}
            \includegraphics[trim={0 3cm 0 3cm},clip,height=1.5em]{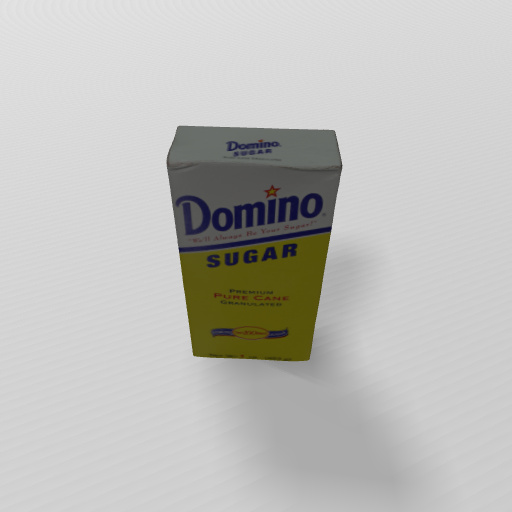} & 004\_sugar\_box & \textbf{\large \;\sfrac{5}{5}} & \textbf{\large \;\sfrac{5}{5}} \\
            \vspace{-0.1cm}
            \includegraphics[trim={0 3cm 0 3cm},clip,height=1.5em]{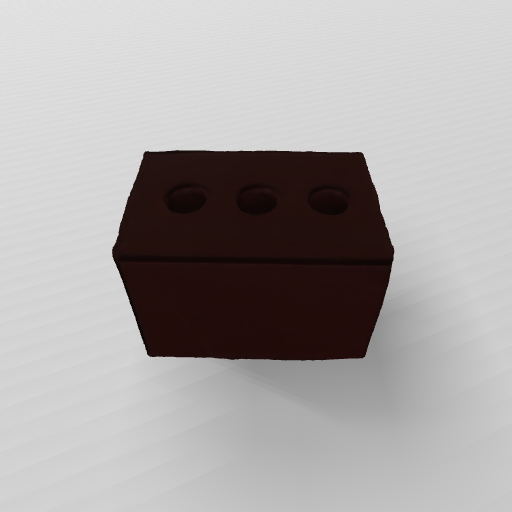} & 061\_foam\_brick & \large \;\sfrac{0}{5} & \textbf{\large \;\sfrac{5}{5}} \\
            \vspace{-0.1cm}
            \includegraphics[trim={0 3cm 0 3cm},clip,height=1.5em]{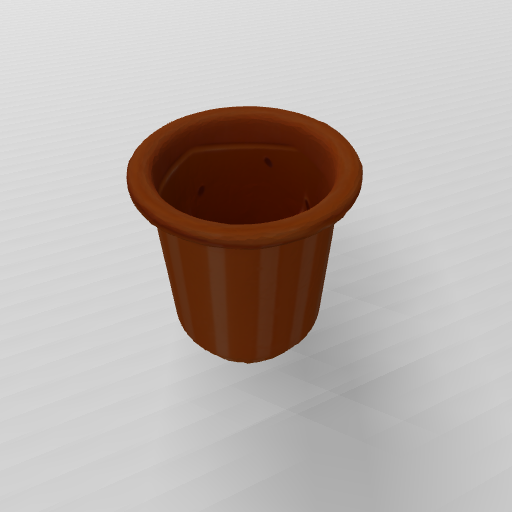} & 065\_cups-a & \large \;\sfrac{0}{5} & \textbf{\large \;\sfrac{4}{5}}  \\ 
            \bottomrule
        \end{tabular}
	}			\\ \vspace*{1ex} {\footnotesize Reported are ~\raisebox{.1mm}{successful\,trials}\,$\sfrac{}{}$\,\raisebox{-.1mm}{number\,of\,trials}.} \vspace*{-2ex}
        \label{tbl:object_wise}
\end{table}

\balance

\section{Discussion and Conclusion}
Our experimental evaluation has shown that the proposed pipeline is able to learn grasping diverse objects in an interactive, human-like manner.
The agent acquires interesting emergent behaviors, such as pre-grasp manipulation and recovery from failures and perturbations.
Further, we have demonstrated how TAPG can adjust learned strategies to a new observation-space, a capability neglected in prior works.
These adjustments resulted in significant improvements to the task's success by obeying the restrictions of the adopted segmentation and tracking model.
Finally, we have shown that our method zero-shot transfers well to the real robot.

An interesting direction for future research is learning to grasp from a deep, cluttered container, where complete occlusions may be unavoidable.

\section*{Acknowledgement}
\small{This work has been funded by the German Ministry of Education and Research (BMBF), grant no. 01IS21080, project “Learn2Grasp: Learning Human-like Interactive Grasping based on Visual and Haptic Feedback”.}

\bibliographystyle{IEEEtran}
\bibliography{references}

\end{document}